\documentclass[journal]{IEEEtran}

\usepackage{booktabs}
\usepackage{multirow}
\usepackage{graphicx}
\usepackage{xcolor}
\usepackage[hyphens]{url}
\usepackage[acronym]{glossaries}
\usepackage[colorlinks=true,allcolors=black]{hyperref} 
\usepackage[sort,compress]{cite}
\usepackage{xspace}
\usepackage{soul} 
\usepackage[utf8]{inputenc}
\usepackage[T1]{fontenc}

\usepackage[caption=false,farskip=0pt]{subfig} 
\usepackage{balance}

\newacronymstyle{long-short-br}
{%
  \GlsUseAcrEntryDispStyle{long-short}%
}%
{%
  \GlsUseAcrStyleDefs{long-short}%
}
\setacronymstyle{long-short-br}

\begin{document}

\title{A New Periocular Dataset Collected by\\Mobile Devices in Unconstrained Scenarios}

\author{Luiz~A.~Zanlorensi,
        Rayson~Laroca,
        Diego~R.~Lucio,
        Lucas~R.~Santos,
        Alceu~S.~Britto~Jr.,
        and~David~Menotti
\thanks{Luiz~A. Zanlorensi, Rayson~Laroca, Diego~R.~Lucio, Lucas~R.~Santos, and David~Menotti are with the Federal University of Paran\'{a} (UFPR), Brazil.
E-mail: \{\textit{lazjunior, rblsantos, drlucio, lrs14, menotti}\}\textit{@inf.ufpr.br}}
\thanks{Alceu~S.~Britto~Jr. is with the Pontifical Catholic University of Paran\'{a} (PUCPR), Brazil.
E-mail: \textit{alceu@ppgia.pucpr.br}}
\thanks{This is an author-prepared version. The final version is published in \textit{Scientific Reports} (DOI: \href{http://doi.org/10.1038/s41598-022-22811-y}{\textcolor{blue}{10.1038/s41598-022-22811-y}}).}
}

\maketitle

\newacronym{cnn}{CNN}{Convolutional Neural Network}
\newacronym{gabor}{GABOR}{Gabor Spectral Decomposition}
\newacronym{hog}{HOG}{Histogram of Oriented Gradients}
\newacronym{lbp}{LBP}{Local Binary Patterns}
\newacronym{nir}{NIR}{near-infrared}
\newacronym{roc}{ROC}{Receiver Operating Characteristic}
\newacronym{safe}{SAFE}{Symmetry Patterns}
\newacronym{sift}{SIFT}{Scale-Invariant Feature Transform}
\newacronym{vis}{VIS}{visible}
\newcommand{\bdcp}{BDCP\xspace}
\newcommand{\crossEyed}{CROSS-EYED\xspace}
\newcommand{\csip}{CSIP\xspace}
\newcommand{\iiitdMSP}{IIITD Multi-spectral Periocular\xspace}
\newcommand{\miche}{MICHE-I\xspace}
\newcommand{\polyu}{PolyU Cross-Spectral\xspace}
\newcommand{\qutMP}{QUT Multispectral Periocular\xspace}
\newcommand{\ubipr}{UBIPr\xspace}
\newcommand{\ubirisvOne}{UBIRIS.v1\xspace}
\newcommand{\ubirisvTwo}{UBIRIS.v2\xspace}
\newcommand{\upol}{UPOL\xspace}
\newcommand{\utiris}{UTIRIS\xspace}
\newcommand{\visob}{VISOB\xspace}
\newcommand{\vssiris}{VSSIRIS\xspace}
\newcommand{\urldataset}{https://web.inf.ufpr.br/vri/databases/ufpr-periocular/}

\begin{abstract}
Recently, ocular biometrics in unconstrained environments using images obtained at visible wavelength have gained the researchers' attention, especially with images captured by mobile devices.
Periocular recognition has been demonstrated to be an alternative when the iris trait is not available due to occlusions or low image resolution.
However, the periocular trait does not have the high uniqueness presented in the iris trait.
Thus, the use of datasets containing many subjects is essential to assess biometric systems' capacity to extract discriminating information from the periocular region.
Also, to address the within-class variability caused by lighting and attributes in the periocular region, it is of paramount importance to use datasets with images of the same subject captured in distinct sessions.
As the datasets available in the literature do not present all these factors, in this work, we present a new periocular dataset containing samples from 1,122 subjects, acquired in 3 sessions by 196 different mobile devices.
The images were captured under unconstrained environments with just a single instruction to the participants: to place their eyes on a region of interest.
We also performed an extensive benchmark with several~\gls*{cnn} architectures and models that have been employed in state-of-the-art approaches based on Multi-class Classification, Multi-task Learning, Pairwise Filters Network, and Siamese Network.
The results achieved in the closed- and open-world protocol, considering the identification and verification tasks, show that this area still needs research and development.
\end{abstract}


\IEEEpeerreviewmaketitle

\section{Introduction}
\label{sec:introduction}

\IEEEPARstart{B}{iometric} systems that use ocular images have been extensively investigated due to the high level of singularity in the iris and because the periocular region can provide discriminative patterns even in noisy images~\cite{santos2013periocular, DeMarsico2017, Proenca2017irina, proenca2019inset, zanlorensi2020attnormalization, zanlorensi2020deep}.
The term ocular comprises the periocular and iris regions~\cite{zanlorensi2019ocular}.
The periocular region comprises eyebrows, eyelashes and eyelids, while the iris is the colored region between the sclera and pupil.
There are two main modes that an ocular biometric system can operate: identification ($1$:$N$ comparison) and verification ($1$:$1$~comparison). 
The identification task consists of determining a subject's identity, whereas the verification one verifies whether a subject is who she/he claims to be.
There are also two main protocols to evaluate biometric systems: closed-world and open-world \cite{zheng2016open, leng2020open}.
In the former, the training and test sets have different samples from exactly the same subjects.
On the other hand, in the open-world protocol, the training and test sets must have samples from different subjects.
With these modes and protocols, it is possible to evaluate some characteristic of biometric approaches to produce discriminative features and generalization capability.

\begin{table}[!ht]
\scriptsize
\setlength{\tabcolsep}{8pt}
\centering
\caption{Comparison of the available ocular datasets containing \gls{vis} images with our dataset (UFPR-Periocular).}

\vspace{-2mm}

\label{tab:datasets}
\resizebox{\linewidth}{!}{
\begin{tabular}{lrrccc}
\toprule
Dataset                                     & Subjects & Images    & Sessions & Sensors \\
\midrule
\vssiris \cite{Raja2015}                     & $28$   & $560$       & $1$      & $2$  \\
\csip \cite{Santos2015}                      & $50$   & $2{,}004$   & N/A      & $7$  \\
QUT~\cite{Algashaam2017}                     & $53$   & $212$       & N/A      & $2$  \\
IIITD \cite{Sharma2014}                      & $62$   & $1{,}240$   & N/A      & $3$  \\
\upol \cite{Dobes2004}                       & $64$   & $384$       & N/A      & $1$  \\
\utiris \cite{Hosseini2010}                  & $79$   & $1{,}540$   & $2$      & $2$  \\
\miche \cite{DeMarsico2015}                  & $92$   & $3{,}732$   & $2$      & $3$  \\
\crossEyed \cite{Sequeira2016, Sequeira2017} & $120$  & $3{,}840$   & N/A      & $2$  \\
\polyu \cite{Nalla2017}                      & $209$  & $12{,}540$  & $2$      & $2$  \\
\ubirisvOne \cite{Proenca2005}               & $241$  & $1{,}877$   & $2$      & $1$  \\
\ubirisvTwo \cite{Proenca2010}               & $261$  & $11{,}102$  & $2$      & $1$  \\
\ubipr \cite{Padole2012}                     & $261$  & $10{,}950$  & $2$      & $1$  \\
\visob \cite{Rattani2016}                    & $550$  & \boldmath{$158{,}136$} & $2$      & $3$  \\
\midrule
\textbf{UFPR-Periocular}                     & \boldmath{$1{,}122$} & $33{,}660$ & \boldmath{$3$} & \boldmath{$196$} \\

\bottomrule

\end{tabular}
}
\end{table}

Nowadays, with the advancement of deep learning-based techniques, several methodologies applying them to ocular images have been proposed for several tasks, for example, spoofing detection~\cite{Menotti2015, He2016}, iris and periocular region detection~\cite{Silva2015, lucio2019simultaneous, severo2018benchmark}, iris and sclera segmentation~\cite{lucio2018fully, bezerra2018robust}, and iris and periocular recognition~\cite{Du2016, Luz2018, Zhao2019capsule, diaz2020spectrum, zanlorensi2018impact, silva2018multimodal, hern2020crossspectral}.
The advancement of these technologies can be observed by the recent contests that have been conducted to evaluate the evolution of the state-of-the-art methods for different applications, such as iris recognition in heterogeneous lighting conditions (NICE.I and NICE.II)~\cite{Proenca2010, Proenca2012}, iris recognition using mobile images (MICHE.I and MICHE.II)~\cite{DeMarsico2015, DeMarsico2017}, iris and periocular recognition in cross-spectral scenarios (Cross-Eyed 1 and 2)~\cite{Sequeira2016,Sequeira2017}, and periocular recognition using mobile images captured in different lighting conditions (VISOB 1 and 2)~\cite{Rattani2016}.
Note that all these contests used datasets containing images obtained in the visible wavelength.
The most recent contests also used images captured by mobile devices~\cite{Rattani2016, DeMarsico2017}.
The results achieved by the proposed methods have shown that it is challenging to develop a robust biometric system in such conditions, mainly due to the high intra-class variability.
Based on recent works~\cite{DeMarsico2017, zanlorensi2019ocular, zanlorensi2020attnormalization}, we can state that developing an ocular biometric system that operates in unconstrained environments is still a challenging task, especially with images obtained by mobile devices.
In this condition, the images captured by the volunteer may present several variations caused by occlusion, pose, eye gaze, off-angle, distance, resolution, and image quality (affected by the mobile~device).

With the existing periocular datasets, it is difficult to assess the scalability performance of biometric applications, i.e., if an approach can produce discriminative features even in a large dataset in terms of the number of subjects.
As we can see in Table~\ref{tab:datasets}, the datasets in the literature do not present a large number of subjects and have few capture devices and session captures.
As described in some previous works~\cite{zanlorensi2020deep, zanlorensi2020attnormalization}, one common problem in ocular biometric systems is the within-class variability, which is generally affected by noises and attributes present in the same individual images.
A robust biometric system must handle images obtained from different capture devices, extracting distinctive representations regardless of the source and environments.
In this sense, samples from the same subject obtained in different sessions are of paramount importance to capture the intra-class variation caused by various noise factors.

\begin{figure}[!ht]
\centering
\begin{tabular}{cccccccc}
    \hspace{-2.7mm}
	{\includegraphics[width=0.11\columnwidth]{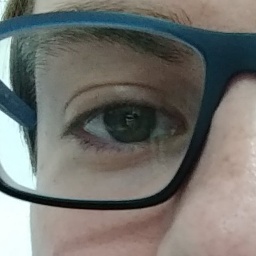}} 
	{\includegraphics[width=0.11\columnwidth]{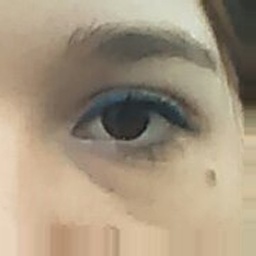}}
    {\includegraphics[width=0.11\columnwidth]{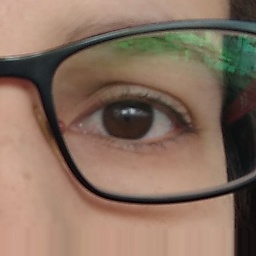}}
    {\includegraphics[width=0.11\columnwidth]{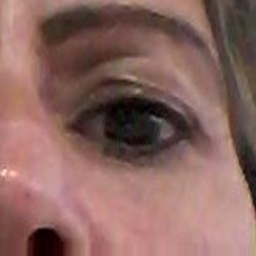}}
    {\includegraphics[width=0.11\columnwidth]{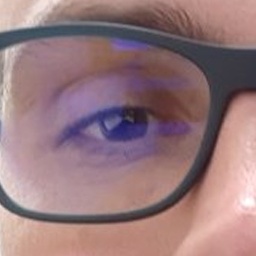}}
    {\includegraphics[width=0.11\columnwidth]{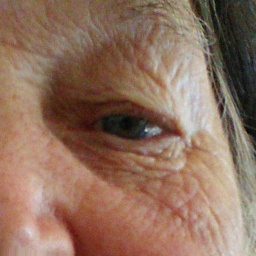}}
    {\includegraphics[width=0.11\columnwidth]{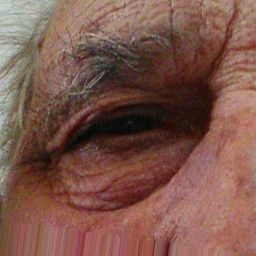}}
    {\includegraphics[width=0.11\columnwidth]{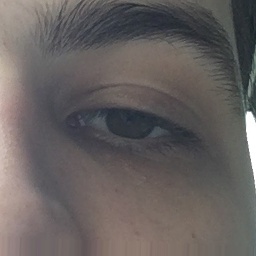}} 
    \\
    \hspace{-2.7mm}
	{\includegraphics[width=0.11\columnwidth]{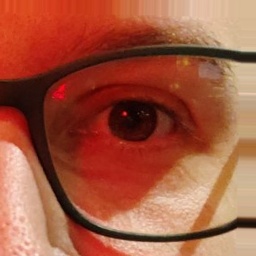}}
	{\includegraphics[width=0.11\columnwidth]{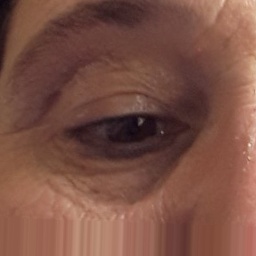}}
    {\includegraphics[width=0.11\columnwidth]{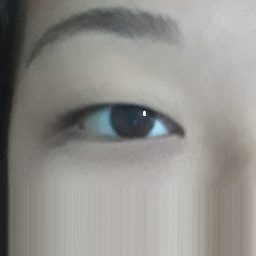}}
    {\includegraphics[width=0.11\columnwidth]{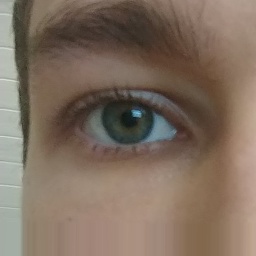}}
    {\includegraphics[width=0.11\columnwidth]{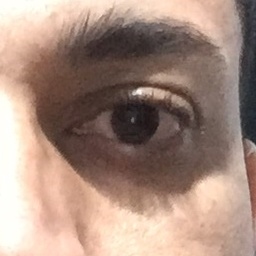}}
    {\includegraphics[width=0.11\columnwidth]{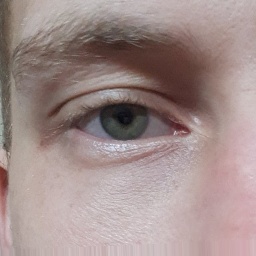}}
    {\includegraphics[width=0.11\columnwidth]{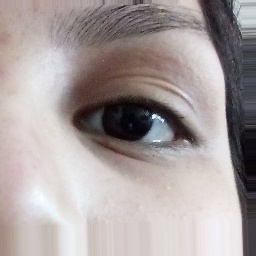}}
    {\includegraphics[width=0.11\columnwidth]{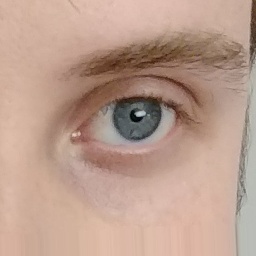}} 
    \\
    \hspace{-2.7mm}
	{\includegraphics[width=0.11\columnwidth]{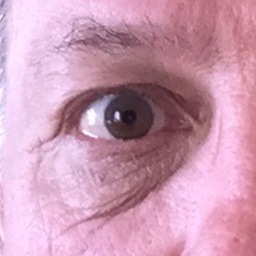}}
	{\includegraphics[width=0.11\columnwidth]{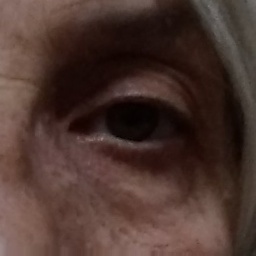}}
    {\includegraphics[width=0.11\columnwidth]{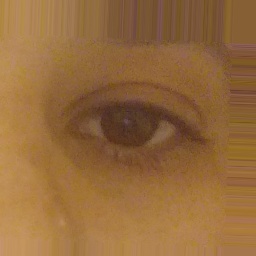}}
    {\includegraphics[width=0.11\columnwidth]{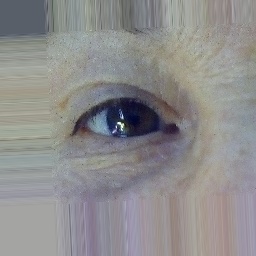}}
    {\includegraphics[width=0.11\columnwidth]{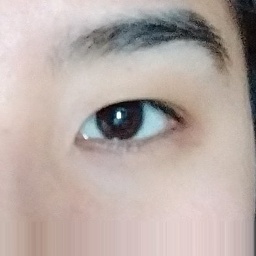}}
    {\includegraphics[width=0.11\columnwidth]{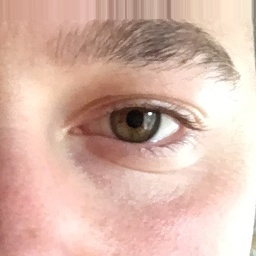}}
    {\includegraphics[width=0.11\columnwidth]{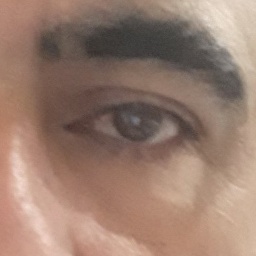}}
    {\includegraphics[width=0.11\columnwidth]{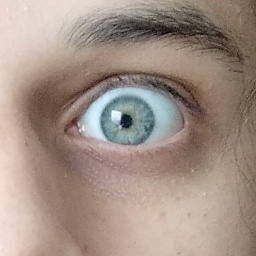}} 
    \\
    \hspace{-2.7mm}
	{\includegraphics[width=0.11\columnwidth]{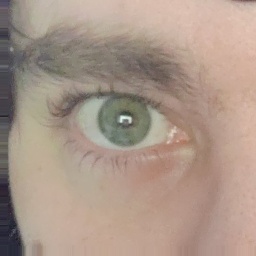}}
	{\includegraphics[width=0.11\columnwidth]{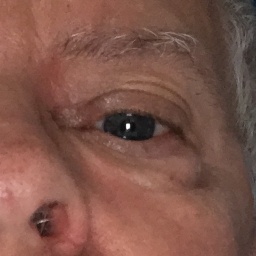}}
    {\includegraphics[width=0.11\columnwidth]{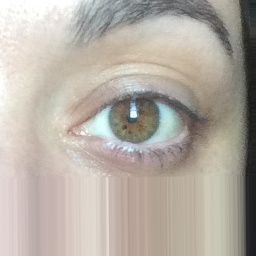}}
    {\includegraphics[width=0.11\columnwidth]{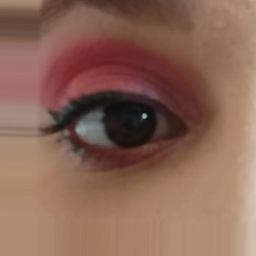}}
    {\includegraphics[width=0.11\columnwidth]{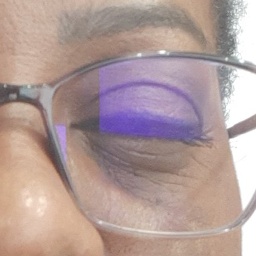}}
    {\includegraphics[width=0.11\columnwidth]{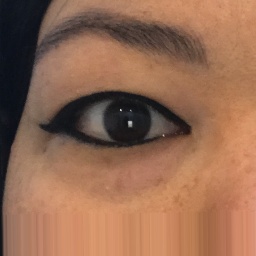}}
    {\includegraphics[width=0.11\columnwidth]{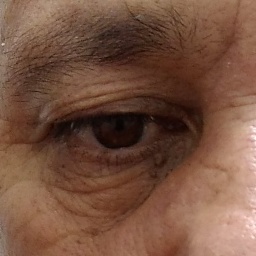}}
    {\includegraphics[width=0.11\columnwidth]{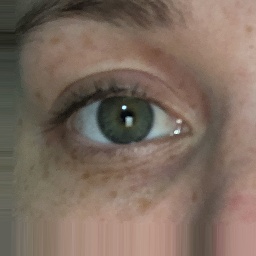}} 
    \\
	\hspace{-2.7mm} {\includegraphics[width=0.11\columnwidth]{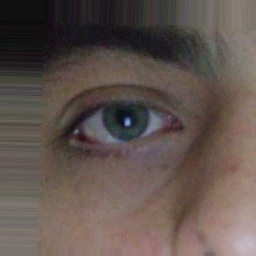}}
	{\includegraphics[width=0.11\columnwidth]{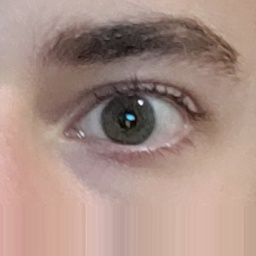}}
    {\includegraphics[width=0.11\columnwidth]{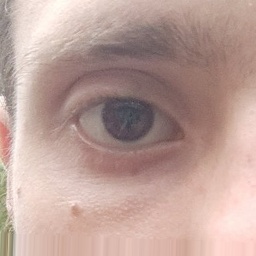}}
    {\includegraphics[width=0.11\columnwidth]{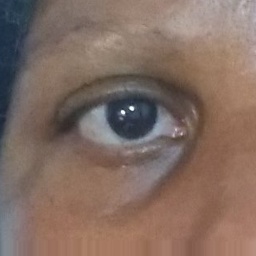}}
    {\includegraphics[width=0.11\columnwidth]{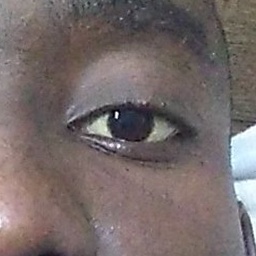}}
    {\includegraphics[width=0.11\columnwidth]{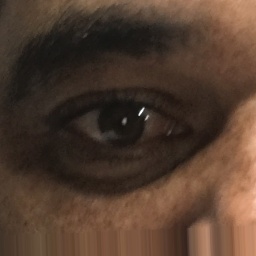}}
    {\includegraphics[width=0.11\columnwidth]{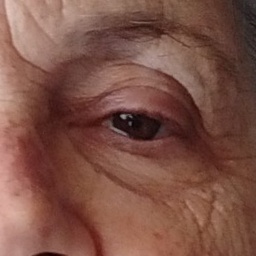}}
    {\includegraphics[width=0.11\columnwidth]{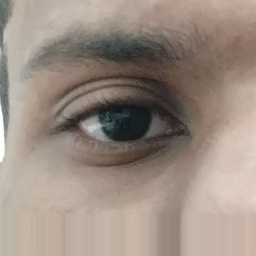}} 
\end{tabular}

\vspace{-1mm}

\caption{Sample images from the UFPR-Periocular dataset. Observe that there is great diversity in terms of lighting conditions, age, gender, eyeglasses, specular reflection, occlusion, resolution, eye gaze, and ethnic diversity.}
\label{fig:datasetsamples}
\end{figure}

Considering the above discussion, in this work, we introduce a new periocular dataset, called \emph{UFPR-Periocular}.
The subjects themselves collected the images that compose our dataset through a mobile application~(app).
In this way, the images were captured in unconstrained environments, with a minimum of cooperation from the participant, and have real noises caused by poor lighting, occlusion, specular reflection, blur, and motion blur.
Fig.~\ref{fig:datasetsamples} shows some samples from the UFPR-Periocular. 
As part of this work, we also present an extensive benchmark, employing several state-of-the-art architectures of~\gls*{cnn} models that have been explored to develop ocular (periocular and iris) recognition biometric~systems.
Face and eye detection are not covered in this work.
The recognition methods are evaluated with manually pre-processed images (also available in the dataset).

Note that our dataset is the largest one in terms of the number of subjects, sessions, and capture devices, as shown in Table~\ref{tab:datasets}.
It also has more images than all datasets except VISOB.
Another key feature is that the proposed dataset has images captured by $196$ different mobile devices.
The samples captured with less cooperation of the participant in unconstrained environments have several variations on the ocular images since they are obtained during three different sessions.
To the best of our knowledge, this is the first periocular dataset with more than $1{,}000$ subject samples and the largest one in different capture devices in the literature.
Thus, we believe that it can provide a new benchmark to evaluate and develop new robust periocular biometric~approaches. 

Recently, with the advancement of devices enabling the self-capture of images that can be used as biometrics, the term ``selfie biometrics'' has been extensively explored by the research community~\cite{rattani2019selfie, selfie2022tapia}, especially in face and iris recognition~\cite{fernandez2019super, kihel2019foveated, arora2019liveness}.
As described by Rattani et al.~[3], the term ``selfie biometrics'' consists of a biometric system where the input data is acquired by the user using the capture devices available in their device.
Thus, we can consider the UFPR-Periocular dataset, presented in this work, as a selfie biometric dataset since its images were acquired by the users through their own~smartphones.

The remainder of this work is organized as follows.
In Section~\ref{sec:related}, we describe the periocular datasets containing~\gls*{vis} images for periocular biometrics.
In Section~\ref{sec:database}, we present information about the UFPR-Periocular dataset and the proposed protocol to evaluate biometric systems.
Section~\ref{sec:benchmark} presents the~\gls*{cnn} architectures used to perform the  benchmark.
In Section~\ref{sec:results}, we present and discuss the benchmark results. 
Finally, the conclusions are given in~Section~\ref{sec:conclusion}.

\section{Related Work}
\label{sec:related}

In recent years, several ocular contests and datasets have been released to evaluate state-of-the-art methods for many applications.
Zanlorensi et al.~\cite{zanlorensi2019ocular} detailed and described several datasets and contests for iris and periocular recognition.
Different problems have been addressed by the researchers, such as ocular recognition in unconstrained environments, ocular recognition on cross-spectral scenarios, iris/periocular region detection, iris/periocular region segmentation, and sclera~segmentation~\cite{vitek2020ssbc}.

Existing periocular datasets can be organized into constrained (or controlled) or unconstrained (or non-controlled) environments.
The quality of the images is different in constrained and unconstrained environments, as some noise can occur in the images captured in unconstrained environments such as lighting variation, occlusion, blur, specular reflection, and distance.
Images can also be acquired cooperatively and non-cooperatively in relation to some image capture restrictions imposed on the subject.
Ocular non-cooperative images can have some problems caused by off-angle, focus, distance, motion blur, and occlusions by some attributes such as eye-glasses, contact lenses, and makeup. 

As described in~\cite{zanlorensi2019ocular}, datasets containing images obtained at the~\gls*{nir} wavelength were created mainly to investigate the intricate patterns present in the iris region~\cite{Phillips2008, Phillips2010}.
There are also other studies on~\gls*{nir} ocular images, such as generating synthetic iris images~\cite{Shah2006, Zuo2007}, spoofing and liveness detection~\cite{Ruiz-Albacete2008, Czajka2013, Gupta2014, Kohli2016}, contact lens detection~\cite{Baker2010, Kohli2013, Doyle2013, Doyle2015}, and template aging~\cite{Fenker2012, Baker2013}.
The use of~\gls*{nir} ocular images captured in controlled environments by biometric systems has been studied for several years.
Thus, it can be considered a mature technology that has been successfully employed in several applications~\cite{bowyer2008survey, Phillips2008, Phillips2010, Proenca2017irina, Proenca2019segmentation}.

In general, better results can be achieved on biometric methods using \gls*{vis} images by exploring the periocular region instead of the iris trait, as the iris is rich in melanin pigment that absorbs the most visible lights --~not reflecting the iris features as occur with \gls*{nir} lights~\cite{bowyer2008survey}.
Also, the small resolution of ocular images is a common problem that makes it almost impracticable to use the iris trait alone.
Regarding these problems, the use of \gls*{vis} ocular images captured in a non-cooperative way under unconstrained environments became a recent challenge. 
In this sense, several studies have been carried out on periocular biometric recognition using images obtained by mobile devices in uncontrolled environments using different capture devices~\cite{DeMarsico2015, Raja2015, Rattani2016}.
The following datasets were developed to investigate the use of iris and periocular traits in \gls*{vis} images: \upol~\cite{Dobes2004}, \ubirisvOne~\cite{Proenca2005}, \ubirisvTwo~\cite{Proenca2010} and \ubipr~\cite{Padole2012}.
There are also datasets of iris and periocular region images for cross-spectral recognition, i.e., match ocular images obtained at different wavelengths (\gls*{nir} against \gls*{vis} and vice-versa): \utiris~\cite{Hosseini2010}, \iiitdMSP~\cite{Sharma2014}, \polyu~\cite{Nalla2017}, \crossEyed~\cite{Sequeira2016, Sequeira2017}, and \qutMP~\cite{Algashaam2017}.
Focusing specifically on ocular recognition using non-cooperative images obtained in uncontrolled environments by mobile devices, we highlight the following datasets: \miche~\cite{DeMarsico2015}, \vssiris~\cite{Raja2015}, \csip~\cite{Santos2015} and~\visob~\cite{Rattani2016}.

Nowadays, it is difficult to evaluate the scalability factor of the state-of-the-art biometric approaches due to the size in terms of subjects and images on the available datasets.
As shown in Table~\ref{tab:datasets}, the most extensive dataset regarding subjects and images is  \visob~\cite{Rattani2016}, which has $158{,}136$ images from $550$ subjects.
The ICIP 2016 Competition on mobile ocular biometric recognition~\cite{Rattani2016} employed this dataset, and in the WCCI/IJCNN2020 challenge (VISOB 2.0 Dataset and Competition results available at \url{https://sce.umkc.edu/research-sites/cibit/dataset.html}), a second version of the dataset was launched.
Both contests evaluated the periocular recognition using \gls*{vis} images obtained by mobile devices.
The second contest's main difference is that the input images were a stack with five periocular images belonging to the same subject.
The best methods achieved an EER of $0.06$\% and $5.26$\% on the first and second contests, respectively.

Also using \gls*{vis} ocular images, other contests were carried out to evaluate iris and periocular recognition: NICE.II~\cite{Proenca2012}, MICHE.II~\cite{DeMarsico2017}, and CROSS-EYED I~\cite{Sequeira2016} and II~\cite{Sequeira2017}.
The NICE.II contest evaluated iris recognition using images containing noise within the iris region.
The winner method fused features extracted from the iris and the periocular region using ordinal measures, color histograms, texton histograms, and semantic information.
The MICHE.II contest also evaluated iris and periocular recognition, but using images captured by mobile devices.
The winner approach extracted features from the iris and the periocular region, using the rubber sheet model normalization~\cite{Daugman1993} and 1-D Log-Gabor filter and Multi-Block Transitional Local Binary Patterns, respectively.
Lastly, the CROSS-EYED I and II contests evaluated iris and periocular recognition on the cross-spectral scenario.
In both contests, the winner approach employed handcrafted features based on~\gls*{safe},~\gls*{gabor},~\gls*{sift},~\gls*{lbp}, and~\gls*{hog}.

Inspired by impressive results achieved by deep learning-based techniques in multiple domains~\cite{lecun2015deep}, several methods proposing and applying such techniques have been developed to address different tasks using ocular images~\cite{Silva2015, lucio2019simultaneous, severo2018benchmark, lucio2018fully, bezerra2018robust, Menotti2015, He2016, Du2016, proenca2019inset, Zhao2019capsule, diaz2020spectrum, zanlorensi2020attnormalization, zanlorensi2018impact, zanlorensi2020deep, Luz2018, silva2018multimodal, hern2020crossspectral}.
Also, as found in the literature, deep learning frameworks for ocular biometric systems are a recent technology that still needs improvement~\cite{zanlorensi2019ocular}.
The use of ocular datasets containing images captured by mobile devices in unconstrained environments is a challenging task that has gained attention in recent years~\cite{DeMarsico2017, Rattani2016, reddy2018comparison, zanlorensi2019ocular, zanlorensi2020attnormalization}.

\section{Dataset}
\label{sec:database}

\begin{table*}[ht]
\centering
\caption{Images, Classes, and Pairwise comparison distributions for the closed-world~(CW) and open-world~(OW) protocols.\hspace{\linewidth}Values for each fold (3 folds).}

\vspace{-2mm}

\label{tab:protocol}
\begin{tabular}{@{}lccccccc@{}}
\toprule
\centering \multirow{3}{*}{Protocol} & \multirow{3}{*}{Train/Val} & \multicolumn{3}{c}{Images / Classes} & \multicolumn{3}{c}{Genuine pairs / Impostor pairs} \\

\cmidrule{3-8}

            & & Train               & Validation         & Test                & Train                        & Validation                 & Test                      \\
\midrule
CW & CW/CW    & $13{,}464/2{,}244$  & $8{,}976/2{,}244$  & $11{,}220/2{,}244$  & $\phantom{0}33{,}660/\phantom{0}90{,}599{,}256$    & $13{,}464/40{,}266{,}336$  & $22{,}440/12{,}583{,}230$ \\
OW & OW/CW    & $13{,}464/1{,}496$  & $8{,}976/1{,}496$  & $11{,}220/\phantom{0,}748$      & $\phantom{0}53{,}856/\phantom{0}90{,}579{,}060$    & $22{,}440/40{,}257{,}360$  & $78{,}540/\phantom{0}4{,}190{,}670$  \\
OW & OW/OW    & $15{,}000/1{,}000$  & $7{,}440/\phantom{0,}496$      & $11{,}220/\phantom{0,}748$      & $105{,}000/112{,}387{,}500$  & $52{,}080/27{,}621{,}000$  & $78{,}540/\phantom{0}4{,}190{,}670$  \\

\bottomrule
\end{tabular}

\end{table*}

\begin{figure}[!ht]
\centering

   	\includegraphics[width=\columnwidth]{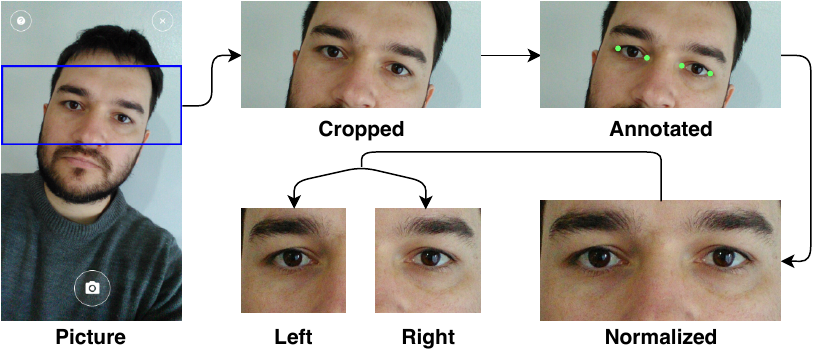}

\vspace{-3mm}

\caption{Image acquisition and normalization process.
First, after the subject took the shot, the rectangular region (outlined in blue) was cropped and stored. 
Then, the images were normalized in terms of rotation and scale using the manual annotations of the corners of the eyes.
Lastly, the normalized images were cropped, generating the periocular regions of the left and right~eyes.}
\label{fig:picprocess}
\end{figure}

The UFPR-Periocular dataset was created to obtain images in unconstrained scenarios that contain realistic noises caused by occlusion, blur, and variations in lighting, distance, and angles.
To this end, we developed a mobile application~(app) enabling the participants to collect their pictures using their smartphones (Project approved by the Ethics Committee Board from the Health Science Sector of the Federal University of Paraná, Brazil -- Process CAAE 02166918.2.0000.0102, registered in the \textit{Plataforma Brazil} system -- \url{https://plataformabrasil.saude.gov.br/}).
We confirm that all methods were carried out following relevant guidelines and regulations by the Ethics Committee Board from the Health Science Sector of the Federal University of Paraná.
Furthermore, we confirm that an informed consent form has been obtained from all subjects, and we do not store any data that could be used to identify the subject.
We confirm that all periocular images presented in this paper (Fig.~\ref{fig:datasetsamples}, Fig.~\ref{fig:picprocess}, Fig.~\ref{fig:multiclass}, Fig.~\ref{fig:multitask}, Fig.~\ref{fig:pairwise}, Fig.~\ref{fig:siamese}, and Fig.~\ref{fig:pairserror}) were extracted from the UFPR-Periocular dataset and that we have permission to publish these images in open access journal.
The single instructions to the participants is to place their eyes on a region of interest marked by a rectangle drawn in the app, as illustrated in ``Picture'' in Fig.~\ref{fig:picprocess}.
We also restricted the images to be captured in $3$ sessions, with $5$ images per session and a minimum interval of $8$ hours between sessions.
In this way, we guarantee that the dataset has samples of the same subject with different noises, mainly due to different lighting and environments.
Furthermore, imposing this minimum time interval between sessions, it is possible to collect different attributes in the periocular region of the same subject, as the images are captured at different times of the day, e.g., subjects wearing and not wearing glasses and makeup.
Another attractive feature of this dataset is that all participants are Brazilian, and as Brazil has great ethnic diversity, there are images of subjects from different races, making this one of the first periocular datasets with such cultural~diversity.

The images were collected from June 2019 to January 2020.
The gender distribution of the subjects is ($53{,}65\%$)~male and ($46{,}35\%$)~female, and approximately $66\%$ of the subjects are under $31$ years old.
In total, the dataset has images captured from $196$ different mobile devices --~the five most used device models were: \textit{Apple iPhone~8}~($4.1$\%), \textit{Apple iPhone~9}~($3.1$\%), \textit{Xiaomi Mi~8 Lite}~($3.0$\%), \textit{Apple iPhone~7}~($3.0$\%), and \textit{Samsung Galaxy J7~Prime}~($2.7$\%).

\begin{figure*}[!ht]
\centering

\resizebox{\linewidth}{!}{
\begin{tabular}{cc}
    
    \includegraphics[height=30ex]{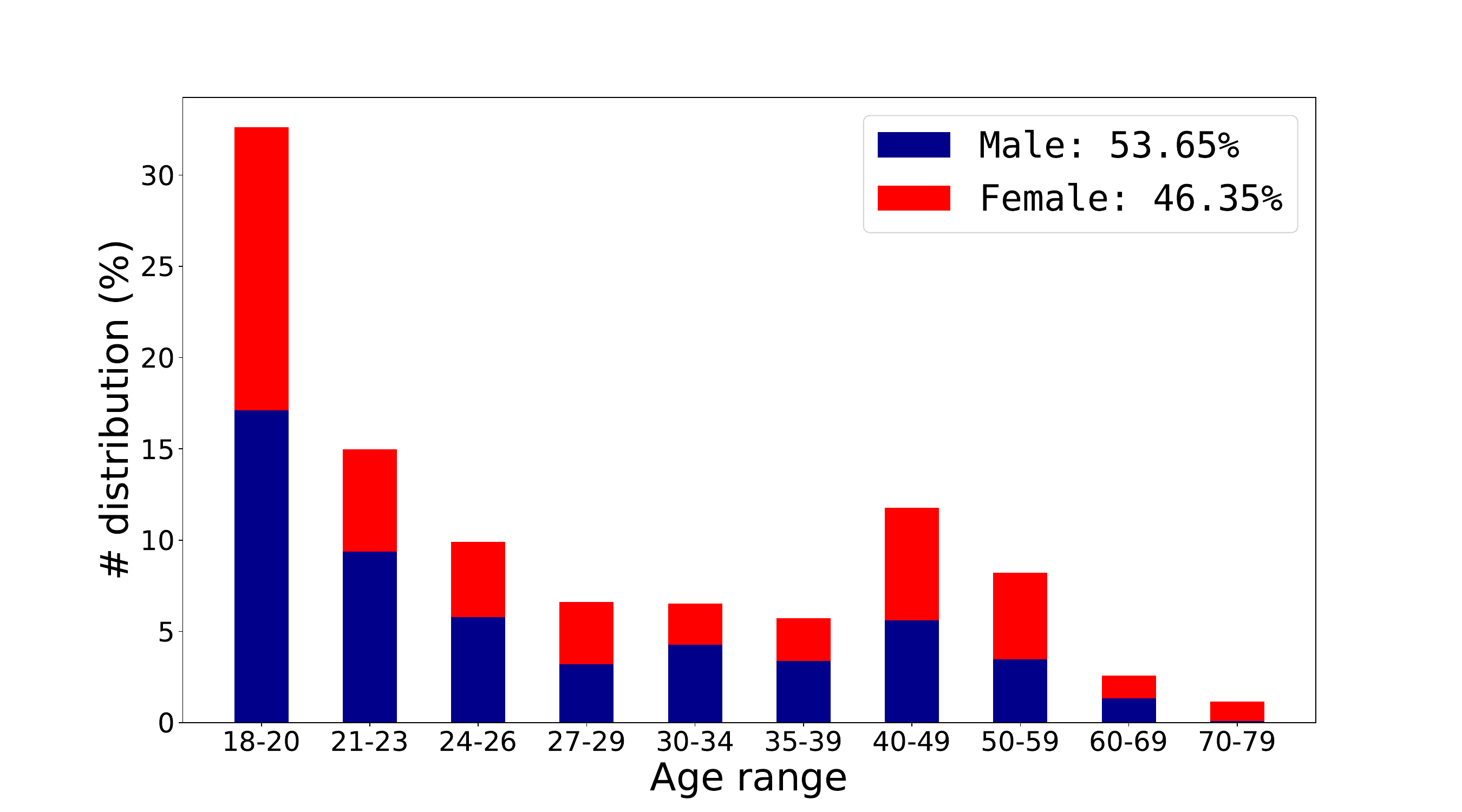}&
    \includegraphics[height=30ex]{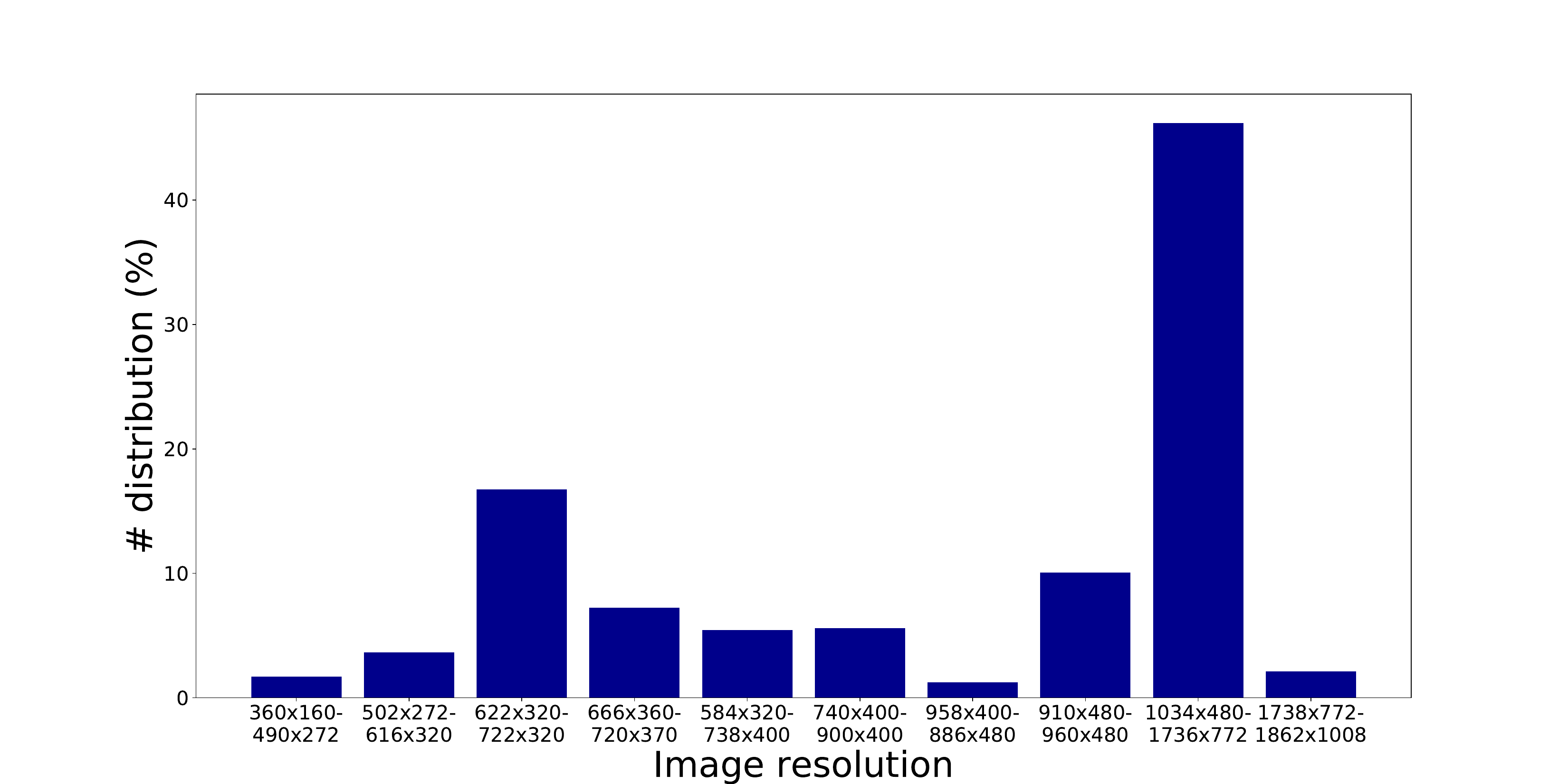} \\[-2pt]

    \footnotesize (a) gender distribution among the age ranges & \footnotesize (b) image resolutions grouped into $10$ intervals
    \normalsize

\end{tabular}
}

\vspace{-1.5mm}

\caption{Age, gender and image resolution distributions in the UFPR-Periocular dataset. (a) note that gender has a balanced distribution, but the age range is concentrated under 30 years old (64\% of the subjects). (b) more than $45$\% of the images have a resolution between $1034\times480$ and $1736\times772$ pixels, and more than $65$\% of the images have resolution higher than $740\times400$ pixels.}
\label{fig:stats}
\end{figure*}

We remark that each subject captured all of their images using the same device model.
The distribution of age, gender, and image resolutions present in our dataset is shown in Fig.~\ref{fig:stats}.

The dataset has $16{,}830$ images of both eyes from $1{,}122$ subjects.
Image resolutions vary from $360\times160$ to $1862\times1008$ pixels --~depending on the mobile device used to capture the image.
We cropped/separated the periocular regions of the right and left eyes to perform the benchmark, assigning a unique class to each side.
Note that, once the image was cropped, the remainder image region was discarded as claimed in our project request to the Ethics Committee Board to preserve at maximum the identity of the participants.
We manually annotated the eye corners with four points per image (inside and outside eye corners) and used these points to normalize the periocular region regarding scale and rotation.
This process is detailed in Fig.~\ref{fig:picprocess}.

Using the center point of each eye (average corners point), the images were rotated and scaled to normalize the eye positions in a size of $512\times256$ pixels.
Then, the images were split into two patches ($256\times256$ pixels) to create the left and right eye sides, generating $33{,}660$ periocular images from $2{,}244$ classes.
The intra- and inter-class variability in this dataset is mainly caused by lighting, occlusion, specular reflection, blur, motion blur, eyeglasses, off-angle, eye-gaze, makeup, and facial~expression.

This new periocular dataset is the main contribution of this work.
It can be employed in future works to evaluate and perform research in biometrics, including recognition, detection and segmentation.
Furthermore, it can also be used to explore studies on recent topics such as gender and age bias~\cite{siddiqui2022bias, ramachandran2022bias}, and to assess the scalability of biometric systems since this dataset is the largest one in the literature in terms of the number of subjects.
Regarding the semantic segmentation problem, we reproduced the experiments presented by Banerjee et al.~\cite{banerjee2022analysis}, which were proposed to generate segmentation masks for iris detection.
This method consists of first transforming the raw image into the HSV and YCbCr color spaces, then using a threshold to binarize both images (HSV mask and YCbCr mask), and finally applying a dot product in both masks to generate the final global mask.
However, as the images from our dataset have considerably more noise than those employed in the original work, the method could not obtain masks of satisfactory quality for us to consider them as ground truth.
For this reason, the semantic segmentation problem will be addressed in future work.

\subsection{Experimental Protocols}
\label{sec:protocol}

We propose protocols for the two most common tasks in biometric systems: identification~($1$:$N$) and verification~($1$:$1$).
The identification task consists of determining a subject sample identity~(probe) within a known dataset or a cluster~(gallery).
The probe is compared against all the gallery samples, considering the closest match as the subject's identity.
Furthermore, probabilistic models can be employed/trained using the gallery data to determine the probe subject's identity based on the highest confidence output.
The verification task refers to the problem of verifying whether a subject is who she/he claims to be.
If two samples match sufficiently, the identity is verified; otherwise, it is rejected~\cite{bowyer2008survey}.
Verification is usually used for positive recognition, where the goal is to prevent multiple people from using the same identity.
The identification is a critical component in negative recognition, where the goal is to prevent a single person from using multiple identities~\cite{jain2008introbiometrics}.
Furthermore, the proposed protocol also encompasses two different scenarios: closed-world and open-world.
In the closed-world protocol, the dataset is split through different samples from the same subject, i.e., training and test sets have samples of the same subjects.
In the open-world protocol, there are different subjects both in the training and test sets.
The identification task is performed in the closed-world protocol, while the verification task can be performed in both closed and open-world protocols.
In the open-world protocol, we also propose two different splits regarding the training and validation sets.
Note that we do not change the test set, keeping it in the open-world protocol, and only vary the training protocols.
The first split uses the closed-world protocol, in which the training and validation sets have samples from the same subjects.
The second split, on the other hand, has different subjects in the training and validation sets, i.e., in an open-world protocol.
With these two training/validation splits, it is possible to use multi-class networks~(classification/identification) and also models based on the similarity of two distinct inputs~(verification task): Siamese networks, triplet networks, and pairwise filters.
Although models built for the verification task can be trained through the closed-world protocol, the design can be better improved using the open-world protocol to split the training and validation sets, as it is a more realistic scenario regarding the test set.
Table~\ref{tab:protocol} summarizes the proposed~protocols.

We defined $3$ folds with a stratified split into training, validation, and test sets for both biometric tasks (identification and verification) for all protocols.
The test set comprises all against all comparisons for genuine pairs and aiming to reduce the pairwise comparisons only impostor pairs using the images of all subjects with the same sequence index, i.e., the $i$-th images of each subject are combined two at-a-time to generate all impostor pairs, for $1\leq i\leq n$, where $n = 3  \text{ sessions} \times 5 \text{ images}$.
As the UFPR-Periocular dataset has images captured under $3$ sessions, we designated one session as a test set for each fold in the \textit{closed-world protocol}.
Thus, we have images from sessions $1$ and $2$, $2$ and $3$, $3$ and $1$ for training/validation, and sessions $3$, $1$, and $2$ for testing, respectively for each of the three folds.  
To evaluate the ability of the models to recognize subjects samples at different environments, for all folds, we employed samples of both sessions in the training and validation sets to fed the models with images from the same subject varying the capture conditions.
For each subject, we employed the first $3$ images of each session for training and the remaining $2$ for validation ($60\%/40\%$ for training/validation splits).
The test set contains new images from the subjects present in the training/validations sets with different noises caused by the environment, lighting, occlusion, and facial attributes.

For the \textit{open-world protocol} we generate the training, validation, and test sets by splitting the dataset through different subjects.
Thus, for each fold, the test set has samples of subjects not present in the training/validation set.
Splitting sequentially by the subject index for each fold, we have samples of $748$ subjects for training/validation and $374$ subjects for testing.
Moreover, we propose two different splits for the training/validation splits, the first one containing images of the same subject in the training and validation sets (closed-world validation).
The second one contains samples from different subjects in the training and validation sets (open-world validation).
Both training/validation protocols have pros and cons.
The advantage of using the closed-world validation is that the training has samples of more subjects than the open-world validation protocol.
However, in this scenario, the models can only learn distinctive features for the gallery samples and may not extract distinctive features for subjects not present in the training process.
On the other hand, the open-world validation has samples of fewer subjects than the closed-world validation protocol, presenting a more realistic scenario since samples of subjects not known in the training stage are present in the validation set.
In the closed-world validation protocol, for each one of the $748$ subjects in the training set, we used the first $3$ images of each session for training, and the remaining $2$ for validation ($60\%/40\%$ for training/validation splits).
In the open-world validation protocol, we employed samples of the first $700$ subjects for training and samples of the remaining $48$ subjects to validate each fold.
The number of the generated pairwise comparison for all protocols are detailed in Table~\ref{tab:protocol}.
The files determining all splits and setups detailed in this section are available along with the UFPR-Periocular~dataset.

\section{Benchmark}
\label{sec:benchmark}

To carry out an extensive benchmark, we employ different models and strategies based on deep learning that achieved promising results in the ImageNet dataset/contest~\cite{deng2009imagenet} and were applied in recent works of ocular recognition~\cite{zanlorensi2018impact, silva2018multimodal, Luz2018, wang2019cross, zanlorensi2020deep}.
These methods differ from each other in network architecture, loss function, and training strategies.
We employed the following \gls*{cnn} models: Multi-class classification, Multi-task learning, Siamese networks, and Pairwise filters networks.
Please note that we did not evaluate detection in this paper.
We employ the images already cropped and resized (to normalize distance and rotation) to evaluate the recognition methods.
In the following subsections, we describe and detail each one of~them.

\subsection{Multi-Class Classification}
\label{sec:multiclass}

\begin{figure}[!ht]
\centering

   	\includegraphics[width=\columnwidth]{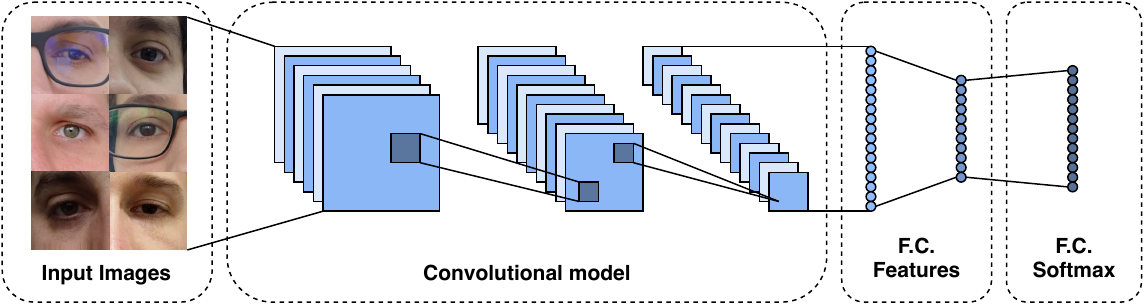}

\vspace{-2mm}

\caption{Multi-class classification \gls*{cnn} architecture.}
\label{fig:multiclass}
\end{figure}

Multi-class classification is the task of classifying instances into three or more classes, where each sample must have a single unique class/label.
Several techniques~\cite{platt1999dags, hastie2000adaboost, huang2012extreme} have been proposed combining multiple binary classifiers to solve multi-class classification problems.
Deep learning-based approaches usually address this problem through \gls*{cnn} models with softmax cross-entropy loss.
Therefore, we start by evaluating several \gls*{cnn} architectures that achieved expressive results in the ImageNet dataset/contest~\cite{deng2009imagenet}.
In summary, the architecture of these models has several convolutional, pooling, activation, and fully-connected layers, as shown in~Fig.~\ref{fig:multiclass}.

In the training stage, a batch of images and their labels feed these models.
The model extracts the image features through convolutional, pooling, and fully connected (dense) layers.
The last layer is composed of a fully connected layer using the softmax cross-entropy as a loss function.
In this work, following previous approaches~\cite{Proenca2010, zhang2016exploring, labati2021social}, we considered each eye of each subject as a unique class, i.e., the left and right eyes belong to different classes.
In this way, as expected, a person's identity can only be verified by the same eye side, i.e., the left and right eyes of the same person can not be matched.
Below we describe the main characteristics of each~model.

\subsubsection{\textbf{VGG}}

The VGG model, proposed by Simonyan and Zisserman~\cite{simonyan2015vgg}, consists of a \gls*{cnn} using small convolution filters ($3\times3$) with a fixed stride of $1$ pixel.
The spatial polling is computed by $5$ max-pooling layers over a $2\times2$ pixel window.
Two models were proposed varying the number of convolutional layers: VGG16 and VGG19.
Both models have two fully connected layers at the top with $4096$ channels each --~these architectures achieved the first and second places in the localization and classification tracks on the ImageNet Challenge 2014.
The authors also stated that it is possible to improve prior-art configurations by increasing the depth of the models.
Parkhi et al.~\cite{parkhi2015vggface} applied these models (called VGG16-Face) on the face recognition problem, showing that a deep \gls*{cnn} with a simpler network architecture can achieve results comparable to the state of the~art.
Furthermore, recent approaches for ocular (iris/periocular) biometrics employing VGG models have demonstrated the ability to produce discriminant features~\cite{zanlorensi2018impact, silva2018multimodal, Luz2018, wang2019cross, zhao2019iriscapsule, zanlorensi2020deep, behera2020twindeep}.
In this work, we employed the VGG16 and VGG16-Face to perform the benchmark.

\subsubsection{\textbf{ResNet}}

The Residual Network (ResNet) was introduced by He et al.~\cite{he2016resnet} and applied to biometrics for face recognition~\cite{cao2017resnetface}, iris recognition~\cite{zanlorensi2018impact, boyd2019deep, zanlorensi2020deep, wang2019cross, zhao2019iriscapsule} and periocular recognition~\cite{zanlorensi2020deep, hern2020crossspectral, behera2020twindeep, boutros2020fusing}.
The authors addressed the degradation (vanishing gradient) problem caused by deeper network architectures proposing a deep residual learning framework.
They added shortcut connections between residual blocks to insert residual information.
These residual blocks are composed of a weighted layer followed by batch normalization, an activation function, another weighted layer, and batch normalization.
Let $F(x)$ be a residual block, and $x$ the input of this block (identity map), the residual information consists of adding $x$ to $F(x)$, i.e., $F(x) + x$, and using it as input to the next residual block.
Different architectures were proposed and evaluated, varying the depth of the models: ResNet50, ResNet101, and ResNet152.
These models achieved promising results on the ImageNet dataset~\cite{deng2009imagenet}.
In~\cite{he2016resnetv2}, He et al. proposed the ResNetV2 by changing the residual block by adding a pre-activation into it.
Empirical experiments showed that the proposed method improved the network generalization ability, reporting better results than ResNetV1~on ImageNet.

\subsubsection{\textbf{InceptionResNet}}

The InceptionResNet model~\cite{szegedy2016inceptionresnet}\st{,} combines the residual connections~\cite{he2016resnet} and the inception architecture~\cite{szegedy2016inception}.
The first inception model~\cite{szegedy2015inception}, known as GoogLeNet, introduced the Inception module aiming to increase the network depth while keeping a relatively low computational cost.
The main idea of inception is to approximate a sparse \gls*{cnn} with a normal dense construction.
The inception module consists of several convolutional layers, where their output filter banks are concatenated and used as the input to the next module.
The model version difference is based on the organization inside its inception module.
Combining the residual connections with the InceptionV3 and InceptionV4 models, the author developed InceptionResNetV1 and InceptionResNetV2, respectively.
Experiments performed on the ImageNet dataset showed that the InceptionResNet models trained faster and reached slightly better results than the inception architecture~\cite{szegedy2016inceptionresnet}.
In our experiments, we employed the InceptionResNetV2 model since it achieved the best results on~ImageNet.

\subsubsection{\textbf{MobileNet}}

The first version of the MobileNet model (MobileNetV1)~\cite{howard2017mobilenet} was developed focusing on mobile and embedded vision applications, in which it is desirable that the \gls*{cnn} model has a small size and high computational efficiency.
This model is based on depthwise separable filters, which are composed of depthwise and pointwise convolutions. 
As described in~\cite{howard2017mobilenet},  depthwise convolutions apply a single filter for each input channel, and pointwise convolutions use a $1\times1$ convolution to compute a linear combination of the depthwise output.
Both layers use batch normalization and ReLU activation.
MobileNetV1 achieved promising results in both terms of performance and accuracy on several tasks such as fine-grained recognition, large scale geolocation, face attributes classification, object detection, and face recognition~\cite{howard2017mobilenet}. 
MobileNetV2~\cite{sandler2018mobilenetv2} combines the first version architecture with an inverted ResNet~\cite{he2016resnet} structure, which has shortcut connections between the bottleneck layers.
Experiments performed in different tasks such as image classification, object detection, and image segmentation showed that the MobileNetV2 can achieve high accuracy with low computation costs compared to state-of-the-art~methods~\cite{sandler2018mobilenetv2}.

\subsubsection{\textbf{DenseNet}}

The Dense Convolutional Network~(DenseNet) model~\cite{huang2017densenet} consists of a \gls*{cnn} architecture where each layer is connected to every other layer in a feed-forward way.
Thus, let $L$ be the number of layers from a network, a DenseNet layer has $\frac{L(L+1)}{2}$ direct connections with subsequent layers --~instead of $L$ as a traditional \gls*{cnn} model.
As in the ResNet models~\cite{he2016resnet, he2016resnetv2}, these connections can handle the vanishing-gradient problem and ensure maximum information flow between layers.
The feed-forward is preserved, passing the output from all layers as an additional input to the subsequent ones in a channel-wise concatenation.
The DenseNet models achieved state-of-the-art accuracies in image classification on the CIFAR10/100 and ImageNet datasets~\cite{deng2009imagenet, huang2017densenet}. 
The authors proposed different models varying the depth of the network.
In our experiments, we employed DenseNet121~(the shallowest~one).

\subsubsection{\textbf{Xception}}

Xception model was inspired by inception modules, being defined as an intermediate step between convolution and depthwise separable convolution operation~\cite{chollet2017xception}.
The proposed architecture replaces the standard inception modules with depthwise separable convolutions and residual connections.
The Xception is similar to InceptionV3 in terms of parameters but outperforms it on the ImageNet~dataset~\cite{deng2009imagenet}.

\subsection{Multi-Task Learning}
\label{sec:multitask}

Multi-task learning improves generalization using the domain information of related tasks as an inductive bias~\cite{caruana1997multitask}.
This architecture learns several tasks using a shared \gls*{cnn} model, where each task can help the generalization of other tasks.
Caruana~\cite{caruana1997multitask} introduced the Multi-task learning concept and evaluated it in different domains, demonstrating that this method can achieve better results than single-task learning models for related tasks.
In deep neural networks, multi-task learning can be performed by two different setups: hard or soft parameter sharing~\cite{ruder2017multideep}.
All the hidden (convolutional) layer weights are shared in the hard parameter sharing, i.e., the model learns a single representation for all tasks.
In this configuration, it is also possible to add specific layers for different tasks~\cite{laroca2021towards}.
On the other hand, each task is processed by a different model in the soft parameter sharing.
Then, the parameters of these models are regularized to encourage similarities among them.

\begin{figure}[!ht]
\centering

   	\includegraphics[width=\columnwidth]{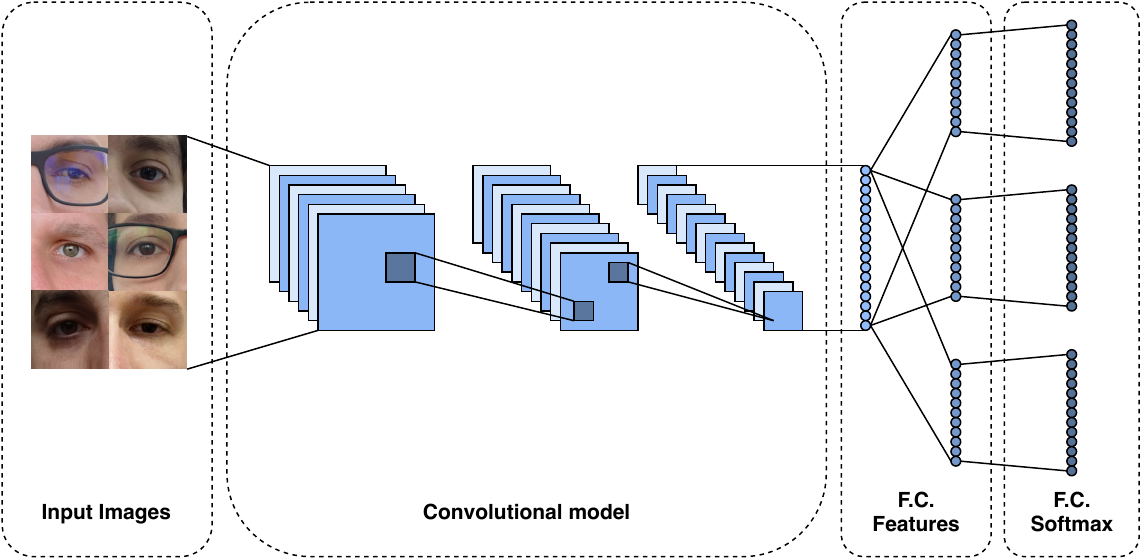}

    \vspace{-2mm}

\caption{Multi-task \gls*{cnn} architecture. In this model, each task has its own output and all tasks share the convolutional layers. The loss of all tasks is used to update the weights of the convolutional layers.}
\label{fig:multitask}
\end{figure}

As shown in Fig.~\ref{fig:multitask}, our Multi-task network shares all convolutional layers and some dense layers.
The model has exclusive dense layers for each task, followed by the prediction layers, using the softmax cross-entropy as function~loss.

In this work, based on the results of multi-class classification, we employ  MobileNetV2 as the base model on our multi-task approach.
Furthermore, as detailed in Table~\ref{tab:multitask-architecture}, we build our multi-task model with hard parameter sharing for the following $5$ tasks: (i)~class prediction, (ii)~age rate, (iii)~gender, (iv)~eye side, and (v)~smartphone model.

\begin{table}[!htb]
\centering
\caption{Multi-task architecture in the closed-world protocol.}
\label{tab:multitask-architecture}

\vspace{-2mm}

\resizebox{.99\columnwidth}{!}{ %
\begin{tabular}{@{}clccc@{}}
\toprule
\textbf{\#} & \textbf{Layer}      & \textbf{Connected to}  & \textbf{Input}             & \textbf{Output} \\ \midrule
$0$  & MobileNetV2 ($88$ layers)   & --                     & $224 \times 224 \times 3$  & $1280$ \\
$1$  & dense (classes)             & $\#0$                  & $1280$                     & $256$ \\
$2$  & dense (age)                 & $\#0$                  & $1280$                     & $256$ \\
$3$  & dense (gender)              & $\#0$                  & $1280$                     & $256$ \\
$4$  & dense (eye side)            & $\#0$                  & $1280$                     & $256$ \\
$5$  & dense (smartphone model)    & $\#0$                  & $1280$                     & $256$ \\
$6$  & predict (classes)           & $\#1$                  & $256$                      & $2244$ \\
$7$  & predict (age)               & $\#2$                  & $256$                      & $10$ \\
$8$  & predict (gender)            & $\#3$                  & $256$                      & $2$ \\
$9$  & predict (eye side)          & $\#4$                  & $256$                      & $2$ \\
$10$ & predict (smartphone model)  & $\#5$                  & $256$                      & $196$ \\ \bottomrule

\end{tabular}} %
\end{table}

For the age estimation task, we generate the classes by grouping ages into the following $10$ ranges: $18$-$20$, $21$-$23$, $24$-$26$, $27$-$29$, $30$-$34$, $35$-$39$, $40$-$49$, $50$-$59$, $60$-$69$, and $70$-$79$.
The gender and eye side prediction tasks have only $2$ classes, while the smartphone model prediction has $196$~classes.
Note that it is possible to employ weighted loss for each task in the Multi-task learning networks, penalizing the wrong classification of some tasks more than others.
For simplicity, in this work, we do not use weighted losses in our experiments, giving equal importance to all tasks.

As shown in Table~\ref{tab:multitask-architecture}, we build exclusive dense layers for each task by connecting them directly to the backbone model (MobileNetV2).
Then, each dense layer is connected to its respective prediction layer, making it possible that each task has its own specialized (feature) dense~layer.

\subsection{Pairwise Filters Network}
\label{sec:pairwise}

\begin{figure}[!ht]
\centering

   	\includegraphics[width=\columnwidth]{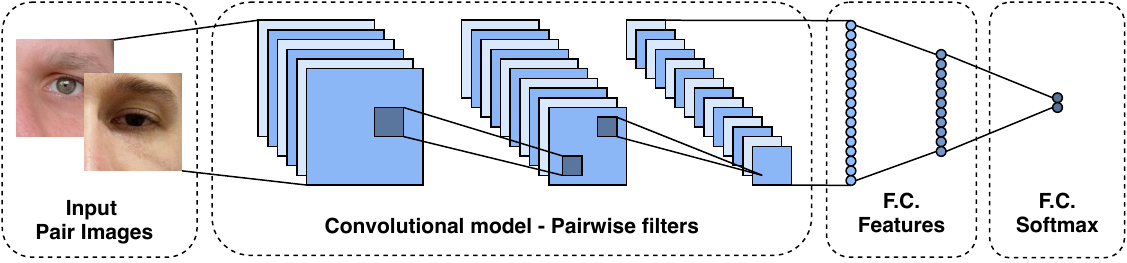}

\vspace{-2mm}

\caption{Pairwise filters \gls*{cnn} architecture. This model contains filters that directly learn the similarity between a pair of images. The output informs whether the images are of the same person or not.}
\label{fig:pairwise}
\end{figure}

Inspired by Liu et al.~\cite{liu2016deepiris}, which is one of the first works applying deep learning for iris verification, we also evaluate the performance of the pairwise filters network.
This kind of model directly learns the similarity between a pair of images through pairwise filters.
The Pairwise Filters Network is a Multi-class classification model that contains one or two outputs informing whether the input pairs are from the same class or from different classes.
The difference is that the network input is a pair of images instead of a single image.
Thus, the network architecture consists of convolutional, pooling, activation, and fully connected layers, as shown in Fig.~\ref{fig:pairwise}.

As described by Liu et al.~\cite{liu2016deepiris}, in this kind of model the similarity map is generated through convolution and summarizes the feature maps of a pair of input images.
We generate the input pairs by concatenating the images at their channel levels.
Let two RGB images with shapes of $224\times224\times3$, concatenating both images by their channels; the resulting input image will have a shape of $224\times224\times6$ ($224\times244$ pixels by 6 channels, 3 from the first image and 3 from the second image).
These images proceed through convolution layers that generate feature maps regarding their similarity.
The output of our model has two neurons and uses a softmax cross-entropy loss.
As the verification problem has only two classes, this model's output can have only one neuron using a binary cross-entropy loss function.
As in the Multi-task network, we employ MobileNetV2 as the base model for our Pairwise Filters~Network.

\subsection{Siamese Network}
\label{sec:siamese}

\begin{figure}[!ht]
\centering
   	\includegraphics[width=\columnwidth]{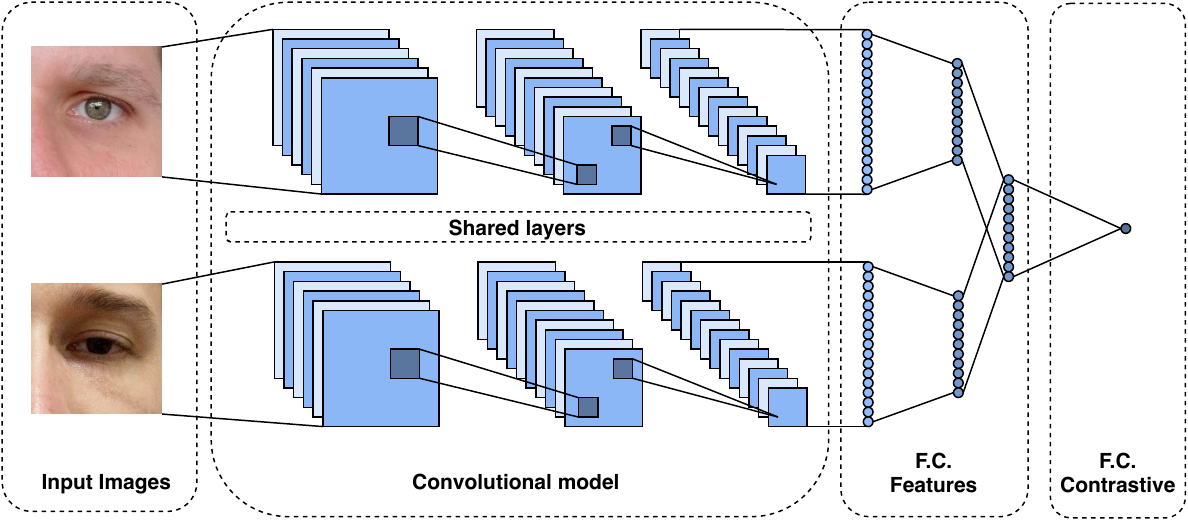}

\vspace{-2mm}

\caption{Siamese \gls*{cnn} architecture. This model is composed of two twin branches of convolutional layers sharing their trainable parameters. The output computes a distance between the input image~pairs.}
\label{fig:siamese}
\end{figure}

Siamese networks were first described by Bromley et al.~\cite{bromley1993siamese} for signature verification.
This architecture consists of twin branches sharing their trainable parameters.
Such models are generally employed for verification tasks since they learn similarities/distances between a pair of inputs.
As illustrated in Fig.~\ref{fig:siamese}, each branch of the Siamese structure is composed of a \gls*{cnn} model followed by some dense layers.
These models can also have shared and non-shared dense layers at the top. 

As detailed in Table~\ref{tab:siamese-architecture}, we employ MobileNetV2 as the base model for each branch of the Siamese network.
We use the contrastive loss~\cite{chopra2005contrastive, hadsell2006contrastive} in the training stage to compute the similarity between the input pair~images. 

\begin{table}[!htb]
\centering
\caption{Siamese network architecture description.}
\label{tab:siamese-architecture}

\vspace{-2mm}

\resizebox{.99\columnwidth}{!}{ %
\begin{tabular}{@{}llccc@{}}
\toprule
\textbf{\#} & \textbf{Layer}  & \textbf{Connected to} & \textbf{Input}            & \textbf{Output} \\ \midrule
$0$ & branch\_a (MobileNetV2 ($88$ layers)) & --                    & $224 \times 224 \times 3$ & $256$ \\
$1$ & branch\_b (MobileNetV2 ($88$ layers)) & --                    & $224 \times 224 \times 3$ & $256$ \\
$2$ & dense              & \#0 and \#1           & $512$                     & $256$ \\
$3$ & Euclidean dist. / Contrastive loss        & \#2                   & $256$                     & $1$ \\ \bottomrule

\end{tabular}
}
\end{table}

As described in~\cite{hadsell2006contrastive}, let $D_W$ be the Euclidean distance between two input vectors, the contrastive loss can be written as follows:

\begin{equation}
C(W) = \sum_{i=1}^{P}L(W,(Y,\vec{X_{1}},\vec{X_{2}})^{i}),
\end{equation}

\noindent where

\begin{equation}
L(W,(Y,\vec{X_{1}},\vec{X_{2}})^{i}) = (1 - Y)L_{S}(D_{W}^{i}) + YL_{D}(D_{W}^{i}) \, ,
\end{equation}

\noindent and $P$ is the number of training pairs, $(Y,\vec{X_{1}},\vec{X_{2}})^{i}$ corresponds to the $i$-th label ($Y$) of the sample pair $\vec{X_{1}},\vec{X_{2}}$, and $L_{S}$ and $L_{D}$ are partial losses for a pair of similar and dissimilar points, respectively.
The objective of this function is to minimize $L$ for $L_{S}$ and $L_{D}$ by computing low and high values of $D_{W}$ for similar and dissimilar pairs, respectively.  

The contrastive loss was proposed and applied to face verification~\cite{chopra2005contrastive, hadsell2006contrastive} and has been employed for periocular recognition~\cite{zhao2018improving, behera2020twin} and iris recognition~\cite{wang2019cross}.

\section{Results and Discussion}
\label{sec:results}

This section presents the benchmark results for the identification and verification tasks.
We first describe the experimental setup used to perform the benchmark.
Then, we report and discuss the results achieved by each~approach.

\subsection{Experimental Setup}

Inspired by several recent works~\cite{Luz2018, zanlorensi2018impact, reddy2018comparison, wang2019cross, boyd2019fine, zanlorensi2020deep, boutros2020fusing, diaz2020spectrum, hern2020crossspectral}, we perform the benchmark employing pre-trained models on ImageNet and also for face recognition~(VGG16-Face and ResNet50-Face).
Afterward, we fine-tuned these models using the UFPR-Periocular dataset.
Similar to recent works on ocular recognition~\cite{Luz2018, zanlorensi2018impact, silva2018multimodal, zanlorensi2019ocular}, we modify all models by adding a fully convolutional layer before the last layer (softmax) to generate a feature vector with a size of $256$ for each image.
The default input size of the models is $224\times224\times3$, except for the InceptionResNet and Xception models, which have an input size of~$299\times299\times3$. 
Note that the input dimensions are different because we are using pre-trained models and therefore our fine-tuning process should follow the original architectures' input size.
In this way, for training and evaluation, the periocular images were resized to fit the input size required for each method, i.e., $299\times299\times3$ for both InceptionResNet and Xception and $244\times244\times3$ for the remaining~models. 

For all methods, the training was performed during $60$ epochs with a learning rate of $10^{-3}$ for the first $15$ epochs and $5\times10^{-4}$ for the remaining epochs using the Stochastic Gradient Descent~(SGD) optimizer.
Then, we used the weights from the epoch that achieves the lower loss in the validation set to perform the~evaluation.

We employ Rank $1$ and Rank $5$ accuracy for the identification task, and the Area Under the Curve~(AUC), Equal Error Rate~(EER), and Decidability~(DEC) metrics for verification.
Furthermore, to generate the verification scores, we compute the cosine distance between the deep representations generated by each \gls*{cnn} model.
As described and applied in several works with state-of-the-art results~\cite{Luz2018, zanlorensi2018impact, zanlorensi2020deep, zanlorensi2020attnormalization}, the cosine distance is computed by the cosine angle between two vectors, being invariant to scalar transformation.
This measure gives more attention to the orientation than to the coefficient of magnitude of the representations, being an interesting metric to compute the similarity between two vectors.
The cosine metric distance is given by:
\begin{equation}
d_{c}(A,B) = 1 - \frac{\sum_{j=1}^{N}A_{j}B_{j}}{\sqrt{\sum_{j=1}^{N}A_{j}^2} \sqrt{\sum_{j=1}^{N}B_{j}^2}} \,,
\end{equation}
\noindent where $A$ and $B$ stand for the feature vectors.

Regarding the models explicitly developed for the verification tasks, i.e., the Siamese and the Pairwise Filters networks, as this task has unbalanced samples of genuine and impostors pairs, selecting the best samples to perform the training is challenging.
Thus, trying to fit the models by feeding them samples as diverse as possible, we employed all genuine pairs and randomly selected the same number from the impostor pairs for each epoch.
Hence, each epoch may have different impostor samples.
However, for a fair comparison, we generated the random impostor pairs only once for each epoch and fold, and used the same samples for training both~models.

The reported results are from five repetitions for each fold, except for the Siamese and Pairwise filter networks, in which we ran only three repetitions due to the high computational cost.
All experiments were performed on a computer with an AMD Ryzen Threadripper $1920$X $3.5$GHz ($4.0$GHz Turbo)~CPU, $64$~GB of RAM and an NVIDIA Quadro RTX~$8000$ GPU~($48$~GB).
All~\gls*{cnn} models were implemented in Python using the Tensorflow ({\small\url{https://www.tensorflow.org/}}) and Keras ({\small\url{https://keras.io/}})~frameworks.

\subsection{Benchmark results}

The results obtained by each approach in the closed-world and open-world protocols are presented in this section.
An ablation study were performed evaluating each task's influence in the identification mode on the Multi-task learning network.
Table~\ref{tab:modelstats} shows the size and the number of trainable parameters of each \gls*{cnn} model used as a benchmark.
This information was extracted from the models employed in the closed-world protocol since they have more neurons on the last layer than the open-world protocol models.
We also report the results achieved by employing the state-of-the-art method that achieved first place in the VISOB 2 competition on mobile ocular biometric recognition~\cite{nguyen2021visob}.
This method~\cite{zanlorensi2020deep} consists of an ensemble of five ResNet-50 models pre-trained for face recognition and fine-tuned using the periocular images of our dataset and employing the same experimental protocol described in this work.

\begin{table}[!ht]
\centering
\caption{Size~(MB) and number of trainable parameters of the \gls*{cnn}~models used in the benchmark.}
\label{tab:modelstats}

\vspace{-2mm}
\resizebox{.7\columnwidth}{!}{
\begin{tabular}{@{}lrr@{}}
\toprule

Model              & Size (MB)       & Trainable parameters   \\

\midrule
VGG16              & $1088$  & $135{,}886{,}084$  \\
VGG16-Face         & $1088$  & $135{,}886{,}084$  \\ 
InceptionResNet    & $445$   & $55{,}246{,}372$  \\
ResNet50V2         & $400$   & $49{,}786{,}436$  \\
ResNet50           & $198$   & $24{,}609{,}284$  \\
ResNet50-Face      & $198$   & $24{,}609{,}284$  \\
Xception           & $176$   & $21{,}908{,}204$  \\
DenseNet121        & $64$    & $7{,}792{,}964$  \\
MobileNetV2        & $26$    & $3{,}128{,}516$  \\
\midrule
Multi-task         & $37$    & $4{,}494{,}230$  \\
\midrule
Siamese            & $21$    & $2{,}551{,}808$  \\
Pairwise           & $20$    & $2{,}349{,}479$  \\
\bottomrule
\end{tabular}}
\end{table}

As can be seen, the benchmark has a great diversity of models with different sizes and parameters due to their difference in structure, depth, concept, and~architectures. 

\subsubsection{Closed-World Protocol}
\label{sec:closed}
\begin{figure}[!ht]
\centering

   	\includegraphics[width=\columnwidth]{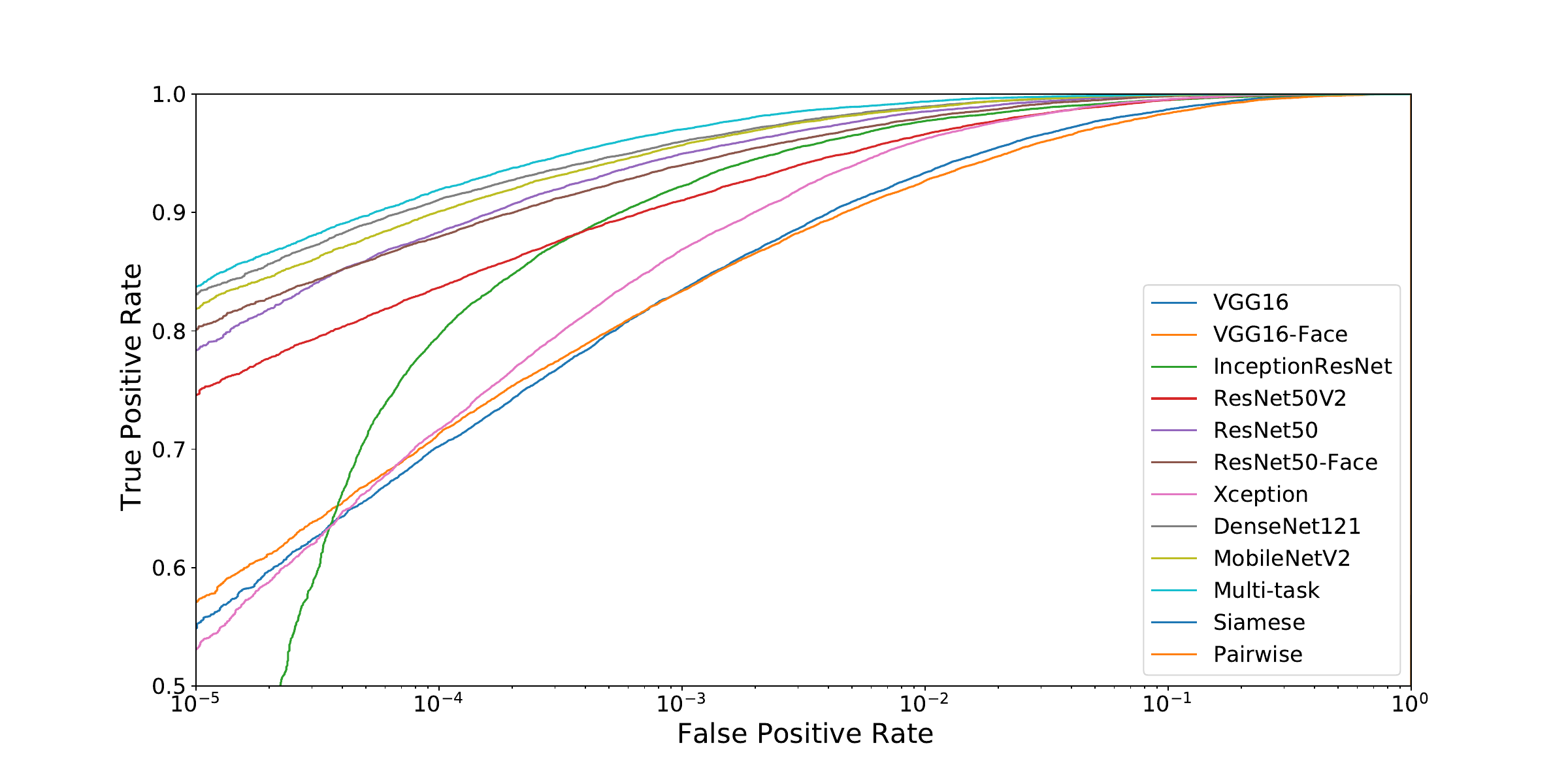}

\vspace{-3mm}
\caption{\acrfull*{roc} curve to compare methods in the closed-world protocol.}
\label{fig:roc_closed}
\end{figure}

We perform the benchmark for both the identification and verification tasks in the closed-world protocol.
All results are presented in Table~\ref{tab:benchclosed} and Fig.~\ref{fig:roc_closed}.
Even though the MobileNetV2 is the shortest model in size and trainable parameters, it achieved the best results for identification and verification tasks.
Therefore, we employed MobileNetV2 as the base model for the Multi-task, Siamese, and Pairwise Filters networks.

\begin{table*}[!ht]
\centering
\caption{Benchmark results in the closed-world protocol for the identification and verification tasks.}
\label{tab:benchclosed}

\vspace{-2mm}
\resizebox{.7\linewidth}{!}{
\begin{tabular}{@{}lccccc@{}}
\toprule

\centering \multirow{2}{*}{Model} & \multicolumn{2}{c}{Identification ($1$:$N$)} & \multicolumn{3}{c}{Verification ($1$:$1$)}  \\

\cmidrule{2-6}

                   & Rank 1 (\%)     & Rank 5 (\%)     & AUC (\%)           & EER (\%)        & Decidability       \\

\midrule
VGG16              & $50.56\pm3.30$  & $68.73\pm3.01$  & $99.41\pm0.11$  & $3.59\pm0.32$   & $4.4544\pm0.1502$ \\
VGG16-Face         & $56.29\pm1.62$  & $73.84\pm1.48$  & $99.43\pm0.08$  & $3.44\pm0.28$   & $4.5069\pm0.1379$ \\
Xception           & $57.43\pm1.43$  & $75.88\pm1.52$  & $99.77\pm0.04$  & $2.19\pm0.18$   & $4.2470\pm0.0538$ \\
ResNet50V2         & $63.18\pm2.14$  & $77.79\pm1.81$  & $99.74\pm0.04$  & $2.24\pm0.18$   & $4.9382\pm0.1184$ \\
InceptionResNet    & $65.16\pm2.45$  & $81.53\pm1.99$  & $99.78\pm0.15$  & $1.85\pm0.40$   & $4.5561\pm0.1183$ \\
ResNet50           & $71.06\pm1.14$  & $85.22\pm0.82$  & $99.89\pm0.02$  & $1.41\pm0.10$   & $5.1242\pm0.0634$ \\
ResNet50-Face      & $73.76\pm1.43$  & $86.86\pm1.02$  & $99.83\pm0.03$  & $1.74\pm0.12$   & $5.2400\pm0.0837$ \\
DenseNet121        & $75.54\pm1.36$  & $88.53\pm0.97$  & $99.93\pm0.02$  & $1.11\pm0.09$   & $5.1730\pm0.0497$ \\
MobileNetV2        & $77.98\pm1.08$  & $90.19\pm0.79$  & $99.93\pm0.01$  & $1.13\pm0.07$   & $5.2477\pm0.0650$ \\
\midrule
\textbf{Multi-task}& $\mathbf{84.32\pm0.71}$  & $\mathbf{94.55\pm0.58}$  & $\mathbf{99.96\pm0.01}$  & $\mathbf{0.81\pm0.06}$   & $5.1978\pm0.0340$ \\
Visob 2.0 Winner~\cite{zanlorensi2020deep, nguyen2021visob} & -  & -  & $99.94\pm0.01$  & $1.02\pm0.09$   & $6.0345\pm0.0788$ \\
\midrule
Siamese            & $-$             & $-$             & $98.94\pm0.22$  & $4.86\pm0.44$  & $3.0005\pm0.1871$  \\
Pairwise           & $-$             & $-$             & $99.44\pm0.66$  & $3.06\pm1.84$  & $\mathbf{6.4503\pm1.2270}$  \\
\bottomrule
\end{tabular}}
\end{table*}

The Multi-task model achieved the best results in Rank~$1$, Rank~$5$, AUC, and~EER metrics.
We emphasize that we only explored other tasks such as --~age, gender, eye side, and mobile device model~-- at the training stage of this model.
We extracted the representations only for the classification task to evaluate identification (using the softmax layer) and verification (using the cosine distance) tasks.
The Siamese network obtained the worst results in the benchmark.
In contrast, the Pairwise Filters network reached the higher Decidability index, indicating that it was the most useful to separate genuine and impostors distributions.
Nevertheless, it did not achieve the best results in terms of AUC and~EER.

The models pre-trained for face recognition generally achieve best results than those pre-trained on the ImageNet~dataset as stated in some previous works~\cite{Luz2018, boyd2019fine}.

\subsubsection{Open-World Protocol}
\label{sec:open}

\begin{figure}[!ht]
\centering

   	\includegraphics[width=\columnwidth]{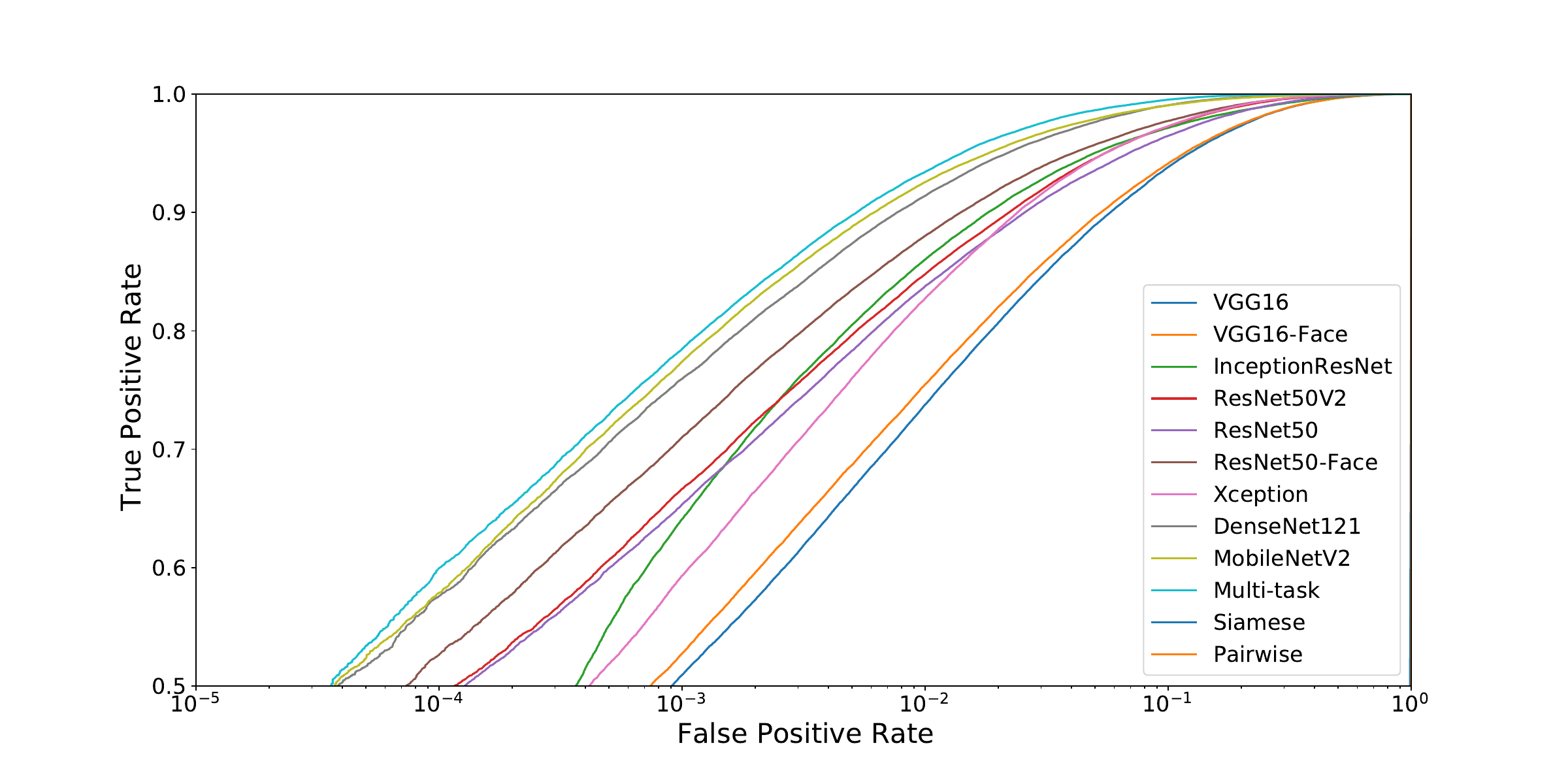}

\vspace{-3mm}
\caption{\gls*{roc} curve to compare methods in the open-world protocol.}
\label{fig:roc_open}
\end{figure}

The main idea of the open-world protocol is to evaluate the capability of the methods to extract discriminant features from samples of classes that are not present in the training stage.
Thus, for this protocol, we perform a benchmark only for the verification task.
The results are shown in Table~\ref{tab:benchopen} and Fig.~\ref{fig:roc_open}.

\begin{table*}[!ht]
\centering
\caption{Benchmark results in the open-world protocol for the verification task.}
\label{tab:benchopen}

\vspace{-2mm}
\resizebox{.6\linewidth}{!}{

\begin{tabular}{@{}lcccc@{}}
\toprule

\centering \multirow{2}{*}{Model} & \centering \multirow{2}{*}{Validation} & \multicolumn{3}{c}{Verification (1:1)} \\

\cmidrule{3-5}

                   &               & AUC (\%)            & EER (\%)        & Decidability       \\

\midrule
VGG16              & Closed-World  & $97.38\pm0.53$  & $8.52\pm0.92$   & $2.9599\pm0.1572$ \\
VGG16-Face         & Closed-World  & $97.70\pm0.42$  & $7.78\pm0.75$   & $3.0327\pm0.1428$ \\
ResNet50           & Closed-World  & $98.60\pm0.28$  & $5.98\pm0.67$   & $3.3702\pm0.1413$ \\
ResNet50V2         & Closed-World  & $98.73\pm0.28$  & $5.69\pm0.64$   & $3.4312\pm0.1459$ \\
Xception           & Closed-World  & $98.93\pm0.16$  & $5.23\pm0.42$   & $3.3493\pm0.0712$ \\
InceptionResNet    & Closed-World  & $99.10\pm0.24$  & $4.61\pm0.65$   & $3.4982\pm0.1208$ \\
ResNet50-Face      & Closed-World  & $99.18\pm0.16$  & $4.38\pm0.47$   & $3.8319\pm0.1239$ \\
DenseNet121        & Closed-World  & $99.51\pm0.12$  & $3.39\pm0.46$   & $3.8646\pm0.1215$ \\
MobileNet          & Closed-World  & $99.56\pm0.08$  & $3.17\pm0.33$   & $3.9868\pm0.1067$ \\
\midrule
\textbf{Multi-task }& \textbf{Closed-World}  & $\mathbf{99.67\pm0.08}$  & $\mathbf{2.81\pm0.39}$   & $3.9263\pm0.0921$ \\
Visob 2.0 Winner~\cite{zanlorensi2020deep, nguyen2021visob}& -  & $99.65\pm0.09$  & $2.96\pm0.26$  & $4.3666\pm0.1453$ \\
\midrule
Siamese            & Closed-World  & $97.27\pm0.64$  & $8.10\pm1.01$   & $2.6678\pm0.2433$ \\
Pairwise           & Closed-World  & $98.62\pm0.72$  & $5.77\pm1.57$   & $\mathbf{4.4404\pm0.5834}$ \\
\midrule
Siamese            & Open-World    & $96.85\pm0.70$  & $8.87\pm1.14$   & $2.6218\pm0.1514$ \\
Pairwise           & Open-World    & $97.80\pm2.03$  & $7.11\pm3.66$   & $4.1977\pm1.0663$ \\
\bottomrule
\end{tabular}}
\end{table*}

As in the closed-world protocol, the Multi-task model achieved the best results in Rank~$1$, Rank~$5$, AUC, and EER, and the Pairwise network achieved the best Decidability index.
The Siamese and Pairwise Filters networks trained using the closed-world validation split reached better results than when trained using the open-world validation split.
We believe this occurred due to the fact that there are fewer classes in the training set in the open-world validation split than in the closed-world validation split.
Although the open-world validation split corresponds to a more realistic scenario regarding the test set, the networks trained with samples from a larger number of classes can reach a higher capability of generalization, producing discriminative representations even for samples from classes not present in the training stage.

\subsubsection{Multi-Task Learning}
\label{sec:multi}

The Multi-task model reached the best results in the closed- and open-world protocols.
As this network simultaneously learns different tasks, we perform an ablation study by running some experiments with $4$ new models created by removing one of the tasks at a time.
The experiments were carried out in the closed-world protocol evaluating the performance for identification and verification.
We also evaluated the results achieved by all models in each~task.

\begin{table*}[!ht]
\centering
\caption{Results (\%) from several Multi-task models trained to predict different tasks. The device model concerns the task of identifying the smartphone model with which the image was taken. The age, gender, and eye side regard the tasks of classifying the input image into age ranges, gender (male or female), and eye side (left or right), respectively.}
\label{tab:multitask}

\vspace{-2mm}
\resizebox{.9\linewidth}{!}{

\begin{tabular}{@{}lccccccc@{}}
\toprule

Model                   & Rank 1     & Rank 5      & Device Model  & Age             & Gender          & Eye Side        \\

\midrule
Multi-task (no model)   & $80.76\pm0.94$  & $91.96\pm0.51$  & $-$               & $82.14\pm0.83$  & $97.72\pm0.17$  & $\mathbf{99.99\pm0.01}$  \\
Multi-task (no age)     & $81.93\pm0.99$  & $93.51\pm0.69$  & $87.20\pm0.63$    & $-$             & $97.65\pm0.20$  & $\mathbf{99.99\pm0.01}$  \\
Multi-task (no gender)  & $82.48\pm0.64$  & $93.55\pm0.52$  & $86.71\pm0.54$    & $83.17\pm0.54$  & $-$             & $\mathbf{99.99\pm0.01}$  \\
Multi-task (no side)    & $83.72\pm0.61$  & $94.07\pm0.54$  & $87.22\pm0.79$    & $83.75\pm0.53$  & $97.70\pm0.20$  & $-$             \\
\textbf{Multi-task}     & $\mathbf{84.32\pm0.71}$  & $\mathbf{94.55\pm0.58}$  & $\mathbf{87.42\pm0.65}$    & $\mathbf{84.34\pm0.71}$  & $\mathbf{97.80\pm0.21}$  & $99.98\pm0.02$  \\

\bottomrule
\end{tabular}}
\end{table*}

According to Table~\ref{tab:multitask}, the Multi-task network without the prediction of the mobile device model was the most penalized for the identification task, followed by the network variations without age, gender, and eye side estimation, respectively.
All models handled the gender and eye side classification tasks well, while the device model and age range classification tasks proved to be more challenging.
One problem in the device model and age range classification is the unbalanced number of samples per class.
Such bias probably contributed to the lower results being achieved in these two~tasks.

Note that we only employed the class prediction for the matching in both closed-world and open-world protocols.
However, as shown in Table~\ref{tab:multitask}, the multi-task architecture also achieved promising results in the other tasks.
In this sense, it may be possible to further improve the recognition results by adopting heuristic rules based on the scores of the other~tasks.

\subsubsection{Subjective evaluation}
\label{sec:subjective}

\begin{figure}[!tb]
\centering
\setlength{\tabcolsep}{1.5pt}
\begin{tabular}{cccc}

    \multicolumn{4}{c}{Wrong Genuines (Best Impostors)} \\
    
    \scriptsize $0.98$ & \scriptsize $0.96$ & \scriptsize $0.95$ & \scriptsize $0.95$\\

    {\includegraphics[width=0.11\columnwidth]{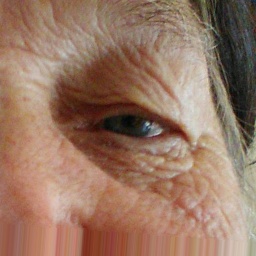}}
    {\includegraphics[width=0.11\columnwidth]{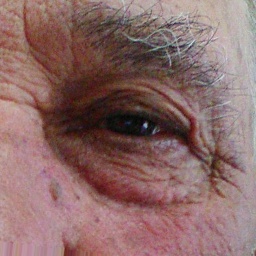}}&
    
    {\includegraphics[width=0.11\columnwidth]{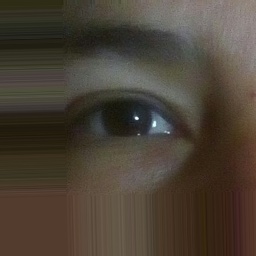}}
    {\includegraphics[width=0.11\columnwidth]{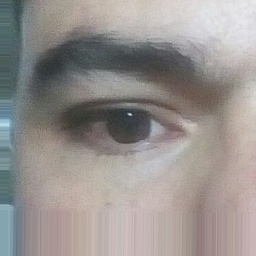}}&
    
    {\includegraphics[width=0.11\columnwidth]{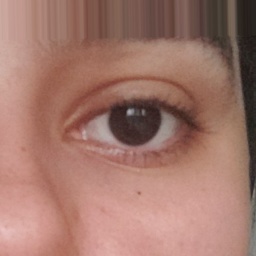}}
    {\includegraphics[width=0.11\columnwidth]{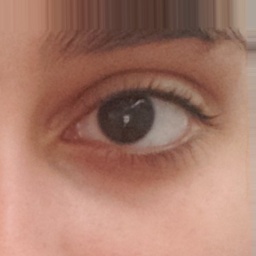}}&
    
    {\includegraphics[width=0.11\columnwidth]{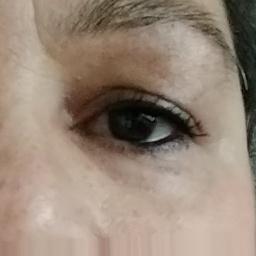}}
    {\includegraphics[width=0.11\columnwidth]{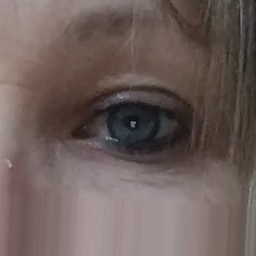}}\\

    \scriptsize $0.95$ & \scriptsize$0.95$ & \scriptsize $0.94$ & \scriptsize $0.94$\\

    {\includegraphics[width=0.11\columnwidth]{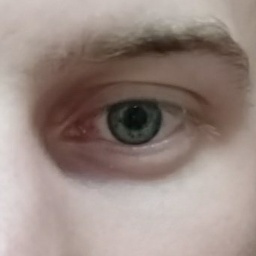}}
    {\includegraphics[width=0.11\columnwidth]{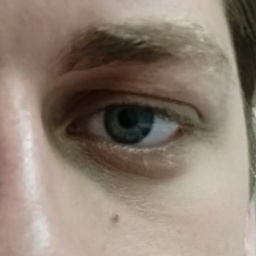}}&
    
    {\includegraphics[width=0.11\columnwidth]{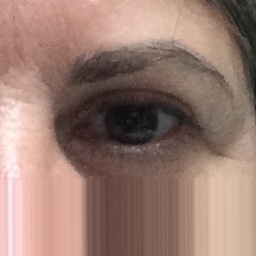}}
    {\includegraphics[width=0.11\columnwidth]{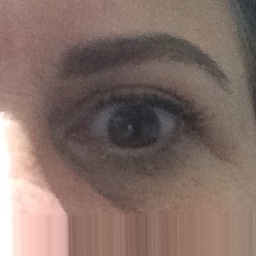}}&
    
    {\includegraphics[width=0.11\columnwidth]{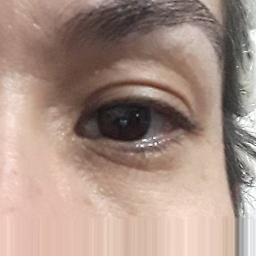}}
    {\includegraphics[width=0.11\columnwidth]{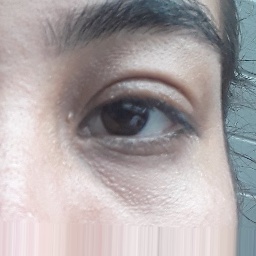}}&
    
    {\includegraphics[width=0.11\columnwidth]{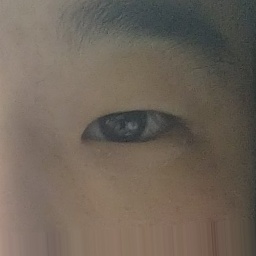}}
    {\includegraphics[width=0.11\columnwidth]{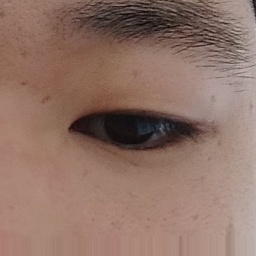}}\\
    
     \scriptsize $0.94$ & \scriptsize $0.94$ & \scriptsize $0.93$ & \scriptsize$0.93$\\

    {\includegraphics[width=0.11\columnwidth]{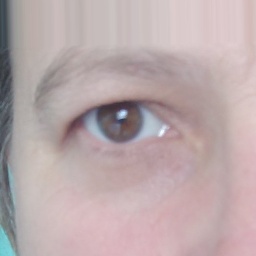}}
    {\includegraphics[width=0.11\columnwidth]{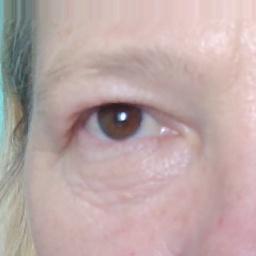}}&

    {\includegraphics[width=0.11\columnwidth]{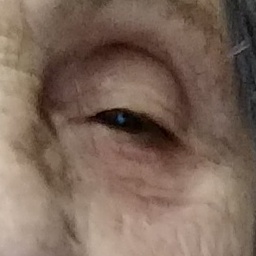}}
    {\includegraphics[width=0.11\columnwidth]{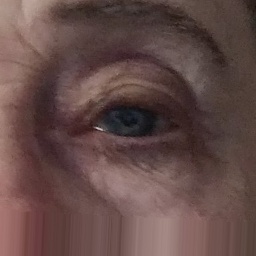}}&

    {\includegraphics[width=0.11\columnwidth]{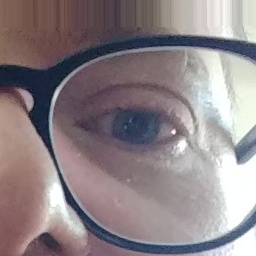}}
    {\includegraphics[width=0.11\columnwidth]{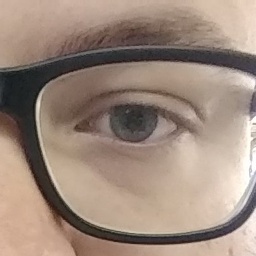}}&
    
    {\includegraphics[width=0.11\columnwidth]{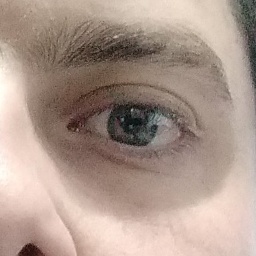}}
    {\includegraphics[width=0.11\columnwidth]{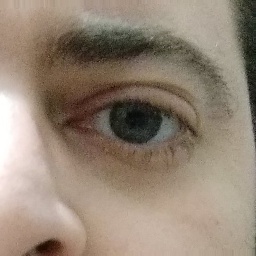}}\\[1.75ex] 
    
    \multicolumn{4}{c}{Wrong Impostors (Worst Genuines)} \\
    
    \scriptsize $0.66$ & \scriptsize $0.68$ & \scriptsize $0.69$ & \scriptsize $0.69$\\

    {\includegraphics[width=0.11\columnwidth]{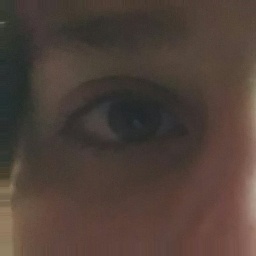}}
    {\includegraphics[width=0.11\columnwidth]{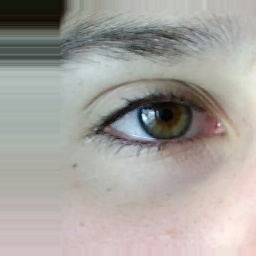}}&

    {\includegraphics[width=0.11\columnwidth]{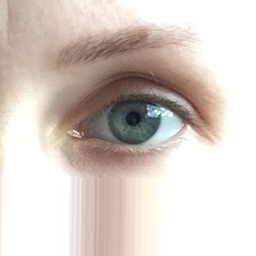}}
    {\includegraphics[width=0.11\columnwidth]{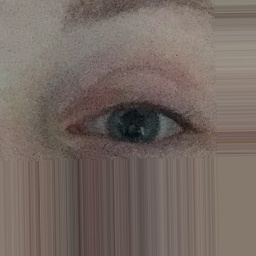}}&

    {\includegraphics[width=0.11\columnwidth]{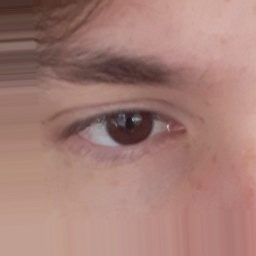}}
    {\includegraphics[width=0.11\columnwidth]{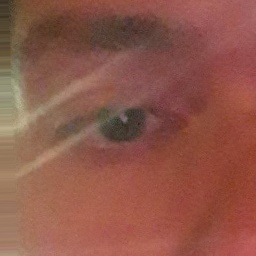}}&
    
    {\includegraphics[width=0.11\columnwidth]{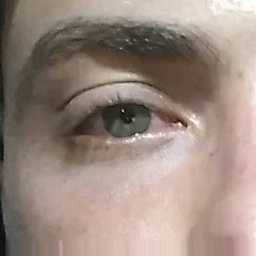}}
    {\includegraphics[width=0.11\columnwidth]{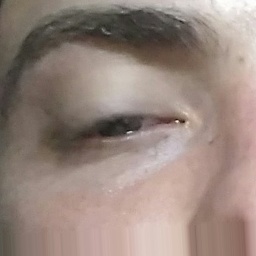}}\\

    \scriptsize $0.69$ & \scriptsize $0.70$ & \scriptsize $0.70$ & \scriptsize $0.70$\\

    {\includegraphics[width=0.11\columnwidth]{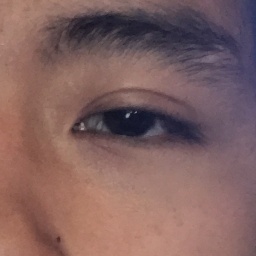}}
    {\includegraphics[width=0.11\columnwidth]{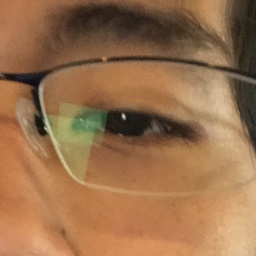}}&

    {\includegraphics[width=0.11\columnwidth]{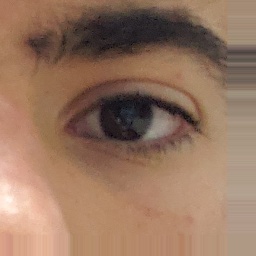}}
    {\includegraphics[width=0.11\columnwidth]{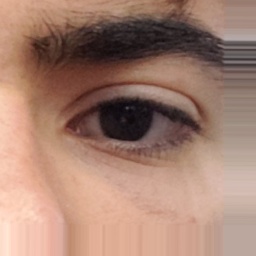}}&

    {\includegraphics[width=0.11\columnwidth]{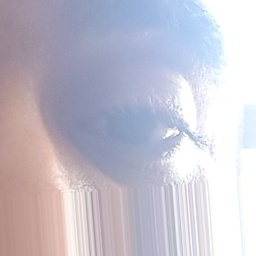}}
    {\includegraphics[width=0.11\columnwidth]{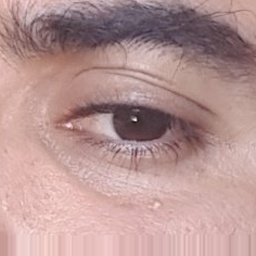}}&
    
    {\includegraphics[width=0.11\columnwidth]{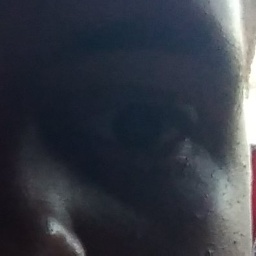}}
    {\includegraphics[width=0.11\columnwidth]{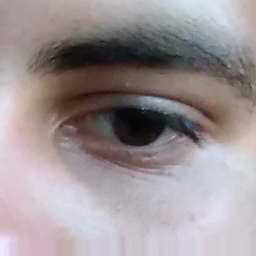}}\\
    
    \scriptsize $0.70$ & \scriptsize $0.71$ & \scriptsize $0.72$ & \scriptsize $0.73$\\

    {\includegraphics[width=0.11\columnwidth]{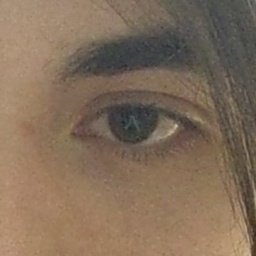}}
    {\includegraphics[width=0.11\columnwidth]{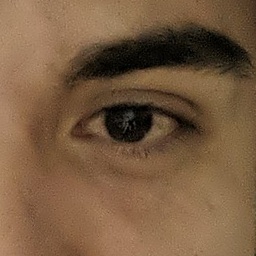}}&

    {\includegraphics[width=0.11\columnwidth]{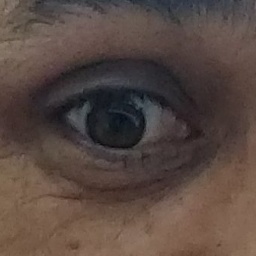}}
    {\includegraphics[width=0.11\columnwidth]{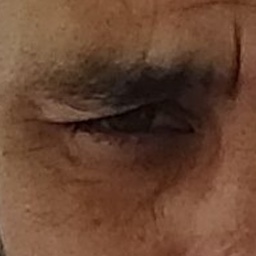}}&

    {\includegraphics[width=0.11\columnwidth]{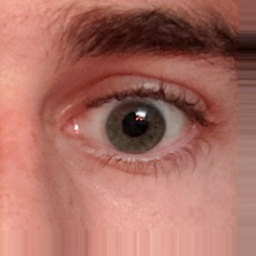}}
    {\includegraphics[width=0.11\columnwidth]{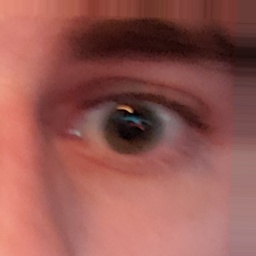}}&
    
    {\includegraphics[width=0.11\columnwidth]{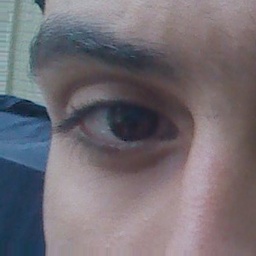}}
    {\includegraphics[width=0.11\columnwidth]{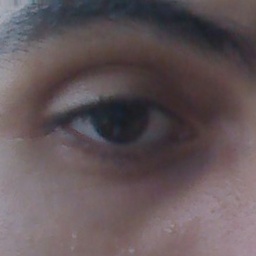}}\\

\end{tabular}
\caption{Pairwise images wrongly classified by the model that obtained the best result in the verification task in the open-world protocol. Higher scores mean that the pair of periocular images is more likely to be~genuine.}
\label{fig:pairserror}
\end{figure}

In this section, we perform a subjective evaluation through visual inspection on the pairs of images erroneously classified by the Multi-task model, which achieved the best result in the verification task in the closed-world protocol.
The best impostors (impostors classified as genuine) and the worst genuines (genuine classified as impostors) pairs are presented in Fig.~\ref{fig:pairserror}.

Performing a visual analysis of all pairwise errors, it is clear that hair occlusion, age, eyeglasses, and eye shape were the most influential factors that led the model to the wrong classification of genuine pairs (intra-class comparison).
In pairs wrongly classified as impostors (inter-class comparison), we saw that lighting, blur, eyeglasses, off-angle, eye-gaze, reflection, and facial expression caused the main difference between the images.
We hypothesize that some errors caused by lightning, blur, reflection, and occlusion can be reduced by employing some data augmentation techniques in the training stage.
Attribute normalization~\cite{zanlorensi2020attnormalization} can also reduce the errors caused by attributes present in the periocular region such as eyeglasses, eye gaze, makeup, and some types of occlusion.
Although some methods can be applied to reduce the matching errors, there are still several characteristics in these images that make the mobile periocular recognition a challenging~task, mainly to the high intra-class variations.

\section{Conclusion}
\label{sec:conclusion}

This article introduces a new periocular dataset that contains images captured in unconstrained environments on different sessions using several mobile device models.
The main idea was to create a dataset with real-world images regarding lighting, noises, and attributes in the periocular region.
To the best of our knowledge, in the literature, this is the first periocular dataset with more than $1{,}000$ subject samples and the largest one in the number of different sensors~($196$).

We presented an extensive benchmark with several~\gls*{cnn} models and architectures employed in recent works for periocular recognition.
These architectures consist of models for multi-class classification and multi-task learning, in addition to Siamese and pairwise filters networks.
We evaluated the methods in the closed-world and open-world protocols, as well as for the identification and verification tasks.
For both protocols and tasks, the multi-task model achieved the best results.
Thus, we conducted an ablation study on this model to understand which tasks significantly influenced the results.
We stated that the mobile device model identification task was the most important, followed by age range, gender, and eye side classification.
Note that we did not conduct experiments employing only left or right eye sides or images separated by gender.
The model trained using all these tasks reported the best result for the identification and verification in the closed- and open-world protocols. 

In a complementary way, we performed a subjective analysis of the best/worst false genuine and true impostors image pairwise comparisons using the Multi-task model, which achieved the best performance for the verification task. 
We observed that lighting, occlusion, and image resolution were the most critical factors that led the model to wrong~verification.

We believe that the UFPR-Periocular dataset will be of great relevance to assist in evolving periocular biometric systems using images obtained by mobile devices in unconstrained scenarios.  
This dataset is the most extensive in terms of the number of subjects in the literature and has natural within-class variability due to samples captured in different sessions.

The Multi-task network using MobileNetV2 as baseline model achieved the best benchmark results for the identification and verification tasks, reaching a rank $1$ of $84,32$\% and an EER of $0.81$\% in the closed-world protocol, and an EER of $2.81$\% in the open-world protocol with thresholds of $0.80$ and $0.78$, respectively.
Therefore, there is still room for improvement in both identification and verification tasks.

\section*{Data availability}

The UFPR-Periocular dataset is publicly available for the research community (upon request) at \href{\urldataset}{\textcolor{blue}{\textit{https://web.inf.ufpr.br/vri/ databases/ufpr-periocular/}}}.
The dataset contains all the original and cropped periocular images, along with the eye corner annotations we made manually.
The files determining all splits and setups for training, validation, and testing employed in our experiments are also part of the dataset as well as information about age, gender, and device model for each image.
We recognize the importance of also providing mask labels for sclera and iris segmentation; however, this is left for future work as making such annotations is time-consuming~\cite{bezerra2018robust,lucio2018fully}.

The UFPR-Periocular dataset is released only to academic researchers from educational or research institutes for non-commercial purposes.
The license agreement must be reviewed and signed by the individual or entity authorized to make legal commitments on behalf of the institution or corporation (e.g., Department/Administrative Head, or~similar).

\section*{Acknowledgments}
This work was supported by grants from the National Council for Scientific and Technological Development~(CNPq) (\#~313423/2017-2 and \#~428333/2016-8) and the Coordination for the Improvement of Higher Education Personnel~(CAPES).
We acknowledge the support of NVIDIA Corporation with the donation of the Quadro RTX $8000$ GPU used for this~research.
\ifCLASSOPTIONcaptionsoff
  \newpage
\fi

\balance

\bibliographystyle{IEEEtran}

\bibliography{references.bib}

\end{document}